\pdfoutput=1
\documentclass[11pt]{article}

\usepackage[]{emnlp2021}

\usepackage{times}
\usepackage{latexsym}

\usepackage[T1]{fontenc}
\usepackage[utf8]{inputenc}

\usepackage{microtype}
\usepackage{graphicx}
\usepackage{amsmath}
\usepackage{amssymb}
\usepackage{booktabs}
\usepackage{bbm}
\usepackage{paralist} % also compact lists
\usepackage{comment}
\usepackage[ruled,vlined]{algorithm2e}
\usepackage{todonotes}
\usepackage{xcolor}
\usepackage{subcaption}
\usepackage{multirow}
\usepackage{makecell}
\usepackage{siunitx}
\usepackage{soul}

\newcommand{\?}{\mbox{?}}

\newcommand{\bi}{\begin{list}{$\bullet$}{
    \setlength{\leftmargin}{1.5 em}
    \setlength{\itemsep}{0 pt}
    \setlength{\topsep}{3 pt}
    \setlength{\parsep}{3 pt}
    \setlength{\partopsep}{0 pt}
    \setlength{\labelwidth}{1 em}
    \setlength{\labelsep}{0.5 em}
    \setlength{\parskip}{0cm}  }}
\newcommand{\ei}{\end{list}}

\newcommand{\BE}{\begin{enumerate}}
\newcommand{\EE}{\end{enumerate}}

\newcommand{\tuple}[1]
        {\mbox{$\langle{#1}\rangle$}}

\newcommand{\initab}{                           % set up tab stops
\begin{tabbing}
XXX \= XXXX \= \kill
}
\newcommand{\begpub}{
\begin{quotation}
\noindent
}

\newcommand{\finpub}{
\end{quotation}
}

\newcommand{\dataset}{\textsc{FloDial}}
\newcommand{\sys}{\textsc{FloNet}}

\newcolumntype{C}[1]{>{\centering\let\newline\\\arraybackslash}m{#1}}
\newcommand{\hlc}[2][yellow]{{%
    \colorlet{foo}{#1}%
    \sethlcolor{foo}\hl{#2}}%
}

\newcommand{\newtext}[1]{\textcolor{black}{#1}}
\newcommand{\newtexta}[1]{\textcolor{black}{#1}}
\newcommand{\textred}[1]{\textcolor{red}{#1}}

\title{End-to-End Learning of Flowchart Grounded Task-Oriented Dialogs}

\author{
Dinesh Raghu$^{\ *\ \dag\ 1\ 2}$, 
Shantanu Agarwal$^{\ *\ 1}$, 
Sachindra Joshi$^{\ 2}$ and
Mausam$^{\ 1}$ \\
$^1$ Indian Institute of Technology, New Delhi, India\\
$^2$ IBM Research, New Delhi, India\\
\texttt{diraghu1@in.ibm.com, a.shantanu08@gmail.com, } \\ 
\texttt{jsachind@in.ibm.com, mausam@cse.iitd.ac.in} \\ \\
\textred{Post-EMNLP Version: Contains Results on the New Hidden Test Set}
}

\begin{document}
\maketitle
\begin{abstract}
%Troubleshooting is an important task handled by customer service agents. During troubleshooting, agents usually follow a flowchart to guide users and help them find a solution. Automating this task is still unexplored. In this paper, we define a novel problem of  learning flowchart grounded task-oriented dialog systems in an end-to-end manner. To investigate this problem, we collect a new dataset called Flowchart Grounded Dialogs (\dataset) dataset. \dataset\ contains a total of 2,738 dialogs grounded on 12 different troubleshooting flowcharts. We propose a baseline named \sys\ for task of flowchart grounded dialog response generation. We evaluate (a) the ability of \sys\ to generate responses by following flowcharts and (b) the ability of \sys\ to generalize to flowcharts unseen during train in a zero-shot flowchart grounded response generation setting. Overall, this paper introduces a new dataset and proposes the first solution to our novel problem setting.

We propose a novel problem within end-to-end learning of task oriented dialogs (TOD), in which the dialog system mimics a troubleshooting agent who helps a user by diagnosing their problem (e.g., car not starting). Such dialogs are grounded in domain-specific flowcharts, which the agent is supposed to follow during the conversation. Our task exposes novel technical challenges for neural TOD, such as grounding an utterance to the flowchart without explicit annotation, referring to additional manual pages when user asks a clarification question, and ability to follow unseen flowcharts at test time.  We release a dataset (\dataset) consisting of 2,738 dialogs grounded on 12 different troubleshooting flowcharts. We also design a neural model, \sys, which uses a retrieval-augmented generation architecture to train the dialog agent. Our experiments find that \sys{} can do zero-shot transfer to unseen flowcharts, and sets a strong baseline for future research.

%We propose a novel problem within end-to-end learning of task oriented dialogs (TOD), in which the dialog system mimics a troubleshooting agent who helps a user by diagnosing their problem (e.g., car not starting). Such dialogs are grounded in domain-specific flowcharts, which the agent is supposed to follow during the conversation. Our task exposes novel technical challenges for neural TOD systems, such as grounding in flowcharts without explicit annotation, referring to additional manual pages when user asks a clarification question, and adaptation to unseen flowcharts at test time.  We release a dataset (\dataset) consisting of 2,738 dialogs grounded on 12 different troubleshooting flowcharts. We also design a novel baseline named \sys\ that uses a retrieval-generation architecture to train the dialog system.
%Our experiments find that FLODIAL outperforms existing models for similar tasks, can do zero-shot transfer to unseen flowcharts, and sets a strong baseline for future research.
%Our experiments find that \sys\ can do zero-shot transfer to unseen flowcharts, and sets a strong baseline for future research.
\end{abstract}

\section{Introduction}

\begingroup
\renewcommand{\thefootnote}{\fnsymbol{footnote}}
\stepcounter{footnote}\footnotetext{D. Raghu and S.Agarwal contributed equally to this work.}
\stepcounter{footnote}\footnotetext{D. Raghu is an employee at IBM Research. This work was carried out as part of PhD research at IIT Delhi.}
%\stepcounter{footnote}\footnotetext{This work was done while S.Agarwal was a research associate at IIT Delhi.}
\endgroup

Task oriented dialog (TOD) systems \cite{BordesW16} converse with users to help them with specific tasks such as calendar enquiry \cite{eric2017key}, restaurant reservation \cite{hen2014word}, and tourist package recommendation \cite{el2017frames}. These dialog systems (e.g., restaurant reservation system) are trained using past human-to-human dialogs and associated knowledge sources (e.g., a KB of restaurants). 
%For example, a restaurant reservation system is trained using past dialogs and a knowledge base of restaurants.

Most existing TOD systems are conversational \emph{recommender} systems %\cite{BordesW16,williams2007partially} 
that gather user requirements in the form of attributes (such as cuisine, location), query a KB and generate recommendations based on the retrieved results (e.g, restaurant, its phone number). While there have been recent efforts \cite{feng2020doc2dial, kim2020beyond} to study non-recommendation TOD,
%beyond conversational recommender systems, plenty of 
several important tasks such as troubleshooting are still unexplored.

\begin{figure*}[ht]
\centering
\includegraphics[width=\textwidth]{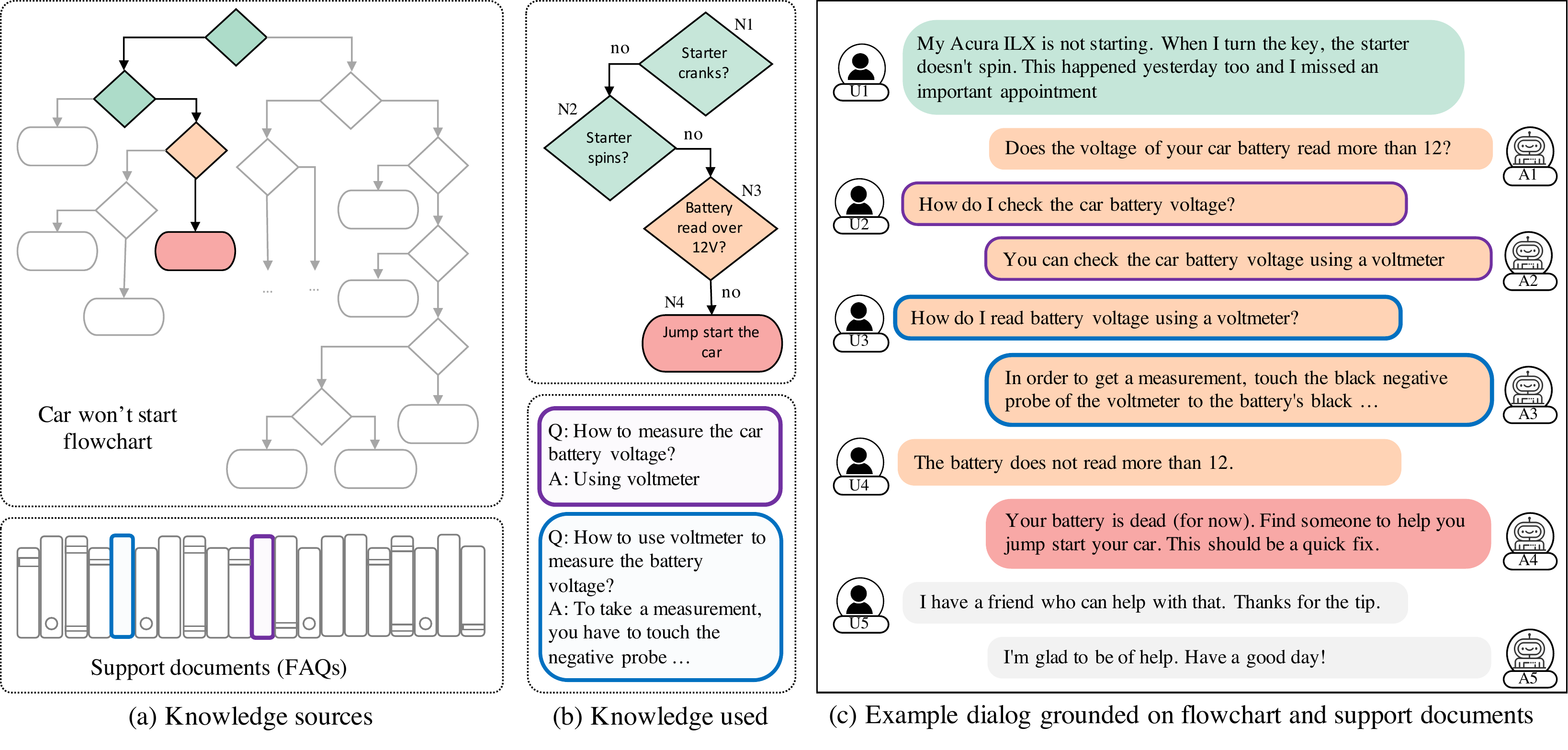}
\caption{Example flowchart grounded TOD. It is grounded on two knowledge sources: flowchart and FAQs.}
\label{fig:intro}
\end{figure*}

Troubleshooting is a common task handled by customer support agents. It involves understanding a user's problem, narrowing down the root cause and providing a solution. Figure \ref{fig:intro} shows an example dialog between an agent and a user troubleshooting a car problem. Support agents typically follow a flowchart (utterances A1, A4 in our example) to diagnose user problems, but may refer to 
supplementary knowledge sources like FAQs (A3), if user asks a clarification question (U3). 
%\todo{May be a good idea to call out turns that are flowchart grounded and the ones which are grounded on FAQs. M:done}

%. The knowledge required for troubleshooting is usually encoded in flowcharts as they crisply summarize pages of unstructured text and enable easy navigation of the overall troubleshooting process. In addition to flowcharts, support agents also use supplementary knowledge sources such as FAQs and support documents.  

In this paper, we propose the novel task of end-to-end learning of a TOD system that troubleshoots user's problems by using a flowchart and a corpus of FAQs.  Our task exposes novel research challenges for TOD system design. First, the system must learn to ground  each utterance in the flowchart without explicit supervision. Second, when required, the agent must refer to additional knowledge in the corpus of FAQs to issue clarifications and add details not present in the flowchart. Third, it must learn the general skill of following a flowchart, which is tested in a zero-shot transfer setting with unseen flowcharts shown at test time. 

%Automating the task of troubleshooting can considerable increase the user satisfaction by reducing their wait time and providing round the clock support. %The primary goal of this paper is to build a task oriented dialog system which can guide users to troubleshoot by following a flowchart and handle digressions using a corpus of FAQs.
%To achieve such automation, we define the novel problem of learning flowchart grounded task oriented dialog system which can guide users to troubleshoot by using a flowchart and a corpus of FAQs. We learn this system in an end-to-end manner without the need for any annotations on the dialogs.
%The labels include (i) whether the response is grounded on a knowledge source, and (ii) the flowchart node (or FAQ) on which the response is grounded on. 

\begin{table}[t]
\centering
\small
\begin{tabular}{l r}
\toprule
\textbf{Utterance Type}  & \textbf{\%} \\
\midrule
{[}T1{]} Problem Description               & 6.09 \\
{[}T2{]}  Grounded on Flowchart             & 56.76 \\
%Informative Digression Question   & 8.57 \\
%Informative Digression Answer     & 6.66 \\
{[}T3{]}  Grounded on Supplementary Knowledge   & 15.23 \\
%Uninformative Digression Question & 3.33 \\
%Uninformative Digression Answer   & 1.90 \\
{[}T4{]}  Chit-Chat                         & 5.23 \\
{[}T5{]}  Conversation Markers              & 7.62 \\
{[}T6{]}  Hold Request                      & 2.38 \\
{[}T7{]}  Reconfirmation                    & 0.95 \\
{[}T8{]}  Dialog Closing                    & 5.71 \\
\bottomrule
\end{tabular}
\caption{Type of utterances and their proportions in a real-world troubleshooting dialog dataset. }
\label{tab:utterance-analysis}
\end{table}

Before collecting a dataset for the task, we first analyze a sample of 100 in-house troubleshooting dialogs with a human customer service agent. Table \ref{tab:utterance-analysis} summarizes the statistics on common utterances in such conversations. This analysis reaffirms the importance of supplementary knowledge (T3). 

We crowdsource the first version of our dataset, \dataset{}\footnote{\url{https://dair-iitd.github.io/FloDial}} ({\bf Flo}wchart Grounded {\bf Dial}ogs),  with these utterance types:  problem description (T1), flowchart following (T2), use of supplementary knowledge in the form of FAQs (T3), and closing utterances (T8). 
%We name this crowdsourced dataset as Flowchart Grounded Dialogs (\dataset). It 
\dataset{} has 2,738 dialogs grounded on 12 different flowcharts.

Since this is a new task, existing end-to-end TOD models are not directly applicable to it. We design a baseline network named \sys\footnote{\url{https://github.com/dair-iitd/FloNet}} -- it
%which takes as input a dialog history, a flowchart, and a corpus of FAQs and predicts the agent response. \sys\ 
follows the retrieval augmented generation framework \cite{ragneurips20} and generates agent response in two steps. First, relevant information from the flowchart and FAQ corpus is retrieved based on the dialog history. Then, this retrieved information and dialog history generate the agent response using an encoder-decoder. We evaluate \sys\ in two different settings: 
%\todo{Can we name them S-Flo and U-Flo instead to indicate seen flowcharts and unseen flowcharts in test.} 
(1) \textit{Seen Flowcharts} (S-Flo) setting, tested on flowcharts seen at train time, and (2) \textit{Unseen Flowcharts} (U-Flo) setting, to evaluate \sys's zero-shot transfer ability in handling new flowcharts unseen at train time. 
%\todo{Shantanu: "unseen" can be replaced by "not seen", sentence sounds better} 
To summarize, the main contributions of this paper are:
\begin{compactenum}
    \item We propose the novel problem of end-to-end learning of flowchart grounded task oriented dialog. 
    %without the use of dialog annotations.
    %We also propose the task of zero-shot flowchart grounded response generation.
    \item We collect a new flowchart grounded task-oriented dialog (\dataset) dataset.
    \item We propose a baseline solution (\sys) for the proposed problem and evaluate it in \textit{seen flowchart} and \textit{unseen flowchart} settings.
\end{compactenum}

%We use the collected dataset to learn a baseline network named \sys, which takes as input a dialog history, a flowchart, and a corpus of FAQs  and predicts the agent response. \sys\ follows the retrieval augmented generation framework \cite{ragneurips20} and generates the response in two steps. In the first step, relevant information from the flowchart and FAQ corpus are retrieved based on the dialog history. Then the retrieved information along with the dialog history is used to generate the agent response. We evaluate \sys\ in two different settings: (1) \textit{in-domain} setting to evaluate \sys's ability to generate responses by following a flowchart where the test dialogs are grounded on flowcharts seen during train, and (2) \textit{out-domain} setting  to evaluate \sys's ability to adapt to zero shot response generation setting where the test dialogs are grounded on flowcharts that were unseen during train.

%Support agents who are trained to troubleshoot using a certain set of flowcharts adapt can troubleshoot using previously unseen flowcharts seamlessly. To study the ability of \sys\ to generalize to a flowchart unseen during train, we evaluate it on a zero-shot flowchart grounded response generation setting.

We release all our resources for further research on the task.

%We will release the \dataset\ dataset and the code for \sys\ for further research.

\section{Related Work}

Dialog systems can be broadly divided into two types: task oriented (TOD) \cite{williams2007partially, BordesW16} and open domain dialog systems \cite{vinyals2015neural,serban2016building}. 
Task oriented dialogs systems can further be divided into end-to-end \cite{BordesW16,raghu-etal-2019-disentangling,raghu-etal-2021-constraint,gangi-reddy-etal-2019-multi} and traditional slot filling approaches \cite{williams2007partially}. Slot filling approaches require dialog state annotations in dialog transcripts. Our work falls under end-to-end approaches, which do not require any such intermediate annotations. 
We first briefly discuss existing TOD datasets and then review approaches for collecting dialog datasets. Finally, we discuss dialog systems related to \sys.

\noindent \textbf{Dialog Datasets:} Exisiting TOD datasets can be grouped based on the type of knowledge source on which the dialogs are grounded. Most of the existing datasets are for the recommendation task and grounded on structured KBs. Some notable KB-grounded datasets are MultiWOZ \cite{budzianowski2018multiwoz}, Stanford multi domain dataset \cite{eric2017key}, CamRest \cite{wenEMNLP2016}, Frames \cite{el2017frames}, schema guided dialogs \cite{rastogi2020towards} and taskmaster-1 \cite{byrne2019taskmaster}. \newcite{kim2020beyond} augment MultiWOZ  with utterances grounded on FAQs. The dialogs in datasets such as ShARC \cite{saeidi2018interpretation} and doc2dial \cite{feng2020doc2dial} are grounded on snippets from unstructured text documents. To the best of our knowledge, \dataset\ is the first TOD dataset that is grounded on flowcharts and FAQs.

%\noindent \textbf{Open Domain Dialog Datasets:} 
%Open domain dialog systems also use external knowledge to generate responses that are interesting, consistent and topically coherent. The CMUDoG dataset \cite{zhou2018dataset} and Wizard-of-Wikipedia \cite{dinan2018wizard} ground their dialogs on Wikipedia articles. The dialogs in OpenDialKG dataset \cite{moon-etal-2019-opendialkg} are grounded on a knowledge graphs which captures the relations between entities and concepts mentioned in the dialogs. While flowcharts and knowledge graphs have a similar structure, the nodes and edges have different semantics associated with them. In the KG used in OpenDialKG, the nodes are entities and edges are relations. Whereas in \dataset the nodes are questions/procedures and edges are the possible user responses.
%\todo{should we add a separating paragraph between Open domain datasets and data collection techniques?}

\noindent \textbf{Dialog Data Collection:} Crowd sourcing frameworks for creating dialog datasets can be broadly grouped into three types. (1) Wizard-of-Oz framework \cite{kelley1984iterative} pairs up two crowd-workers who play the roles of user and agent while conversing. 
%to generate dialogs. One worker plays the role of a user and the other an agent. 
The user is provided with a goal and the agent is given the knowledge necessary to achieve the goal. (2) Self-dialogs framework \cite{byrne2019taskmaster} requires a single crowd-worker to write the entire dialog by playing both user and agent. 
(3) Dialog paraphrasing framework \cite{shah2018building} systematically generates a dialog outline (user and agent utterance) and crowdsources paraphrases for each utterance to construct a dialog. We follow this framework for collecting \dataset, as it gives us adequate control over dialog flow so that we can incorporate various utterance types in Table \ref{tab:utterance-analysis}.

%the dialog paraphrasing framework to collect our \dataset\ dataset.

\noindent \textbf{Dialog Systems:} Large scale pre-trained language models such as GPT2 \cite{radford2019language} have been used for response generation in both open domain \cite{wolf2019transfertransfo,zhang2020dialogpt,zhao-etal-2020-knowledge-grounded} and TOD systems \cite{ham-etal-2020-end,NEURIPS2020_e9462095}. A major  challenge is GPT2's limitation on the input size. For our setting, it becomes difficult to feed a long input (flowchart, dialog history, FAQ corpus)
 to GPT2. We overcome this by following the retrieval augment generation paradigm \cite{ragneurips20} --  we are probably the first to apply it to a dialog setting.

The task of zero-shot response generation requires a model to generalize to new domains with just domain descriptors and no training dialogs. Existing approaches \cite{zhao2018zero,wu2019transferable,rastogi2020towards} model slots and intents as domain descriptors. We model flowcharts as domain descriptors and expect the system to generalize to new flowcharts unseen during train.

\begin{figure*}[ht]
\centering
\includegraphics[width=\textwidth]{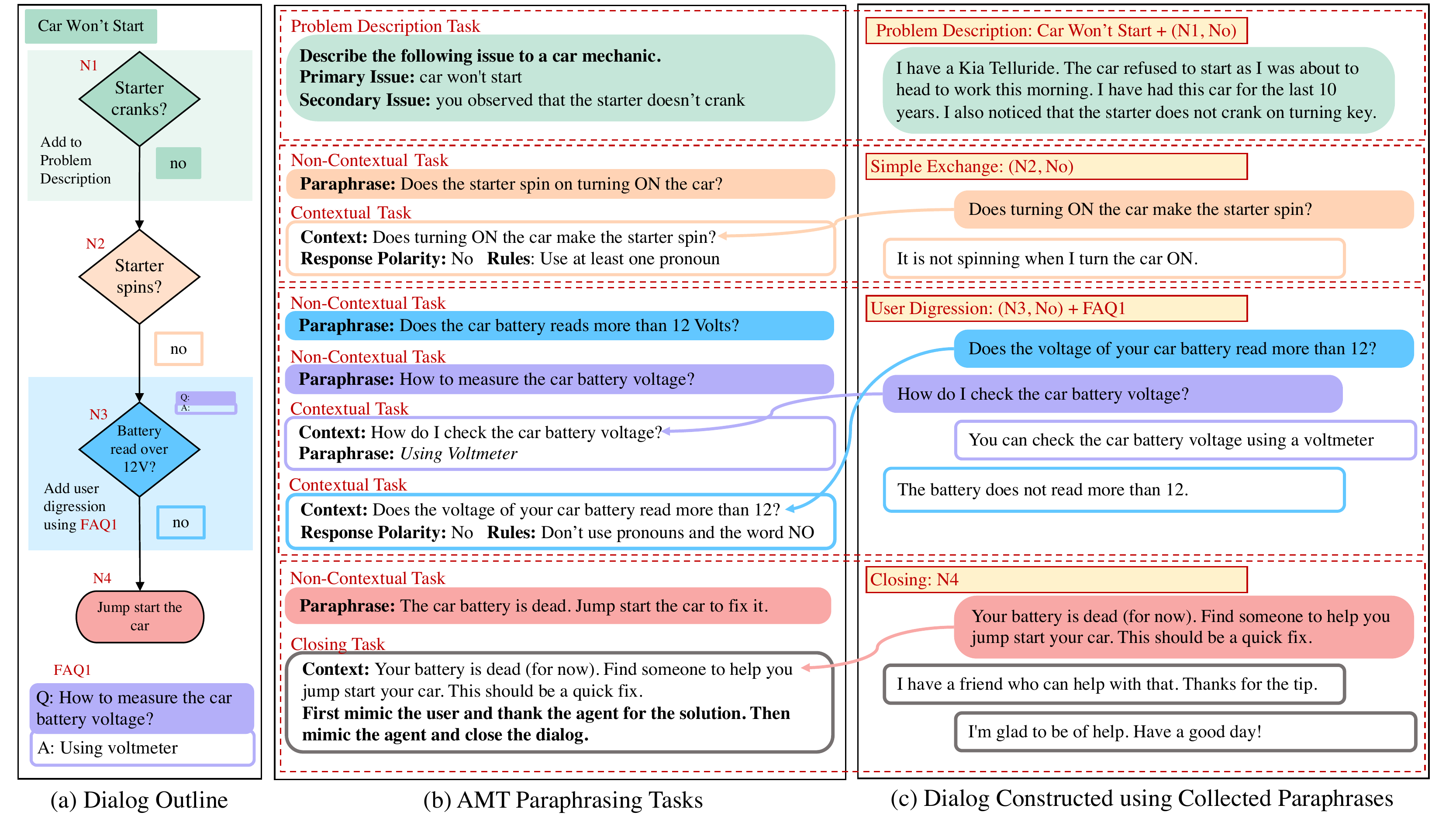}
\caption{An example of a dialog outline, AMT task creation and paraphrasing. Each dotted line box denotes a single component of the dialog. These components are independently paraphrased and then finally stitched together to construct one dialog. Each colored bubble in (b) denotes an AMT task and the matching bubble in (c) denotes the corresponding collected paraphrase. The last two paraphrases in (c) are for the closing task in (b). Paraphrases from non-contextual tasks are used in the corresponding contextual tasks, as denoted by the arrows.}
\label{fig:data-collection}
\end{figure*}

\section{The \dataset\ Dataset}
\label{dataset}
% introduce the dataset, summarize the overall process.

\dataset\ is a corpus of troubleshooting dialogs between a user and an agent collected using Amazon Mechanical Turk (AMT). The dataset is accompanied with two knowledge sources over which the dialogs are grounded: (1) a set of troubleshooting flowcharts and (2) a set of FAQs which contains supplementary information about the domain not present in the flowchart --  both are in English.
%The the collected dataset and the accompanying knowledge sources are both in English. \todo{SA:added dataset language}

%Each agent utterance in the collected dialogs is annotated with the flowchart node/FAQ over which it is grounded.

%In this section, we describe our data collection process based on the dialog paraphrasing framework \cite{shah2018building}. The overall process is illustrated in Figure \ref{fig:data-collection}. 

The data collection process uses the dialog paraphrasing framework \cite{shah2018building} and is illustrated in Figure \ref{fig:data-collection}. 
At a high level, we first systematically construct an outline for each dialog, then decompose the outline into multiple AMT paraphrasing tasks, and finally stitch the dialog using the collected paraphrases. Our data collection process has the following advantages: (i) systematic outline construction guarantees coverage of all paths in the flowchart, and the desired balance of utterance types in dialogs, (ii) the process ensures the annotated labels\footnote{We do not use these annotated labels during train, but use them to evaluate the performance of the dialog system.} are always correct and (iii) it provides diversity in the paraphrases collected.

%In order to propose a new Task-Oriented Dialog challenge for learning flowchart following troubleshooting dialogs, we collect a novel flowchart grounded dataset named \dataset. The dataset consists of dialogs that are based on complex troubleshooting processes related to fixing a car or a laptop. The dialogs were crowdsourced using Amazon Mechanical Turk (AMT), and manually cleaned to ensure high data quality. In the following sections, we describe the dataset and the dialog collection process.

\subsection{Flowcharts and FAQs}

We identify 12 flowcharts\footnote{Downloaded with permission from \url{www.ifitjams.com}} on troubleshooting laptop and cars problems, such as  overheating laptop, car won't start and car brake failure. The flowcharts encode agent questions as decision nodes and user responses as edges. The agent follows flowcharts based on user responses to reach a terminal node (e.g., node N4 in Figure \ref{fig:intro}b) which contains the solution. We refer to the sequence of nodes and edges from root to a terminal node as a path in the flowchart. One such path is shown in Figure \ref{fig:intro}b.

The flowcharts usually contains precise instructions with no details. For example, node N4 in Figure \ref{fig:intro}b just says ``does the battery read over 12V?" but does not provide details such as ``which instrument is needed to measure the battery voltage?" or ``how does one measure the battery voltage using a voltmeter?". For each flowchart, we collect supplementary FAQs\footnote{Collected in-house, refer Appendix \ref{appendix:FAQ}} that contain details such as step-by-step instructions for a process (e.g., ``how to jump start your car?") and other common doubts (e.g., ``where is the ignition coil located?"). A few example FAQs are shown in Figure \ref{fig:intro}b.

% Dialog Outline - how the outline was created - problem statement, collect data for each node-edge pair, propose the solution and finally close the dialog with thank you.

\subsection{Dialog Outline Construction}
\label{section:dialog_construction}
We systematically iterate over paths in the flowchart and for each path we construct multiple outlines. Each outline consists of 3 major parts: problem description, flowchart path traversal and closing. We now discuss each part in detail.

%The dialog outline is a crisp summary of each utterance in the dialog. The outlines were constructed based on insights from in-house customer support troubleshooting dialogs. We analyzed 100 dialogs and identified the following utterance types: problem description (6\%), grounded on flowchart (57\%), grounded on secondary knowledge sources (15\%), closing utterances (6\%) and others (16\%). The dialog outline starts with user's problem description, followed by utterances that are grounded on flowcharts and FAQs. Finally the outline ends with utterances that lead to closing the dialog. 

\vspace{0.5ex} 
\noindent \textbf{Problem Description:} The problem description is the first utterance in a dialog. It contains (1) the primary issue faced by the user, (2) secondary information, and (3) other information that may not be relevant for troubleshooting. The primary issue is phrased using title of the flowchart. For example, for Figure \ref{fig:data-collection} it will be  \textit{car won't start}. The secondary information is any other information that may help in troubleshooting the primary issue. For example, the user could say that \textit{starter is not cranking}. This secondary information is populated by sampling a random (node, edge) pair from the sampled flowchart path. For example, (N1, no) is populated as a secondary information in Figure \ref{fig:data-collection}. 
%We randomly choose to add secondary information to the problem description. This enables us to 
By adding this to problem description, we mimic the setting where, an agent may need to \emph{skip} a few nodes when following the flowchart, based on information already present in the dialog history. 
%the secondary information provided by the user.
%When these observations are mentioned in the problem description, the agent would jump to a relevant node in the flowchart. For example, the agent directed jumped to node \textit{N3} based on the observations mentioned in utterance \textit{U1}.

\vspace{0.5ex} 
\noindent \textbf{Flowchart Path Traversal:} After the problem description, we consider each (node, edge) pair in the given flowchart path. For each (node, edge) pair we toss a coin to decide if the pair should be represented as a simple exchange or as a complex exchange. A simple exchange is one where the agent asks the question in the node and the user responds with the answer in the edge. ($N2$, \textit{No}) in Figure \ref{fig:data-collection}c is constructed as a simple exchange. Complex exchanges use at least four utterances to represent the information in the (node, edge) pair, e.g., ($N3$, \textit{No}) in Figure \ref{fig:data-collection}c. 
%is constructed as complex exchanges. 
Complex exchange can be of two types: user-initiated digression and agent digression. The example illustrates \textit{user digression} where the user asks for clarifications to understand the agent question before responding with an answer. 
%to the question in the node. 
An \textit{agent digression} is similar except that the agent proactively breaks a complex question into a sequence of simple ones. 
%\todo{Not clear by what is meant by agent digression. Example needed. - sachin}
An example agent digression for ($N3$, \textit{No}) would be when the agent first asks ``Do you know how to measure the voltage of a car battery using a voltmeter\?". If the user responds ``no", the agent will then describe the procedure to measure the voltage, and then requests the user to check if the voltage is  greater than 12V.

%(2) When the agent breaks down a question in a single node into a series of question to avoid miscommunication. This usually happens when the agent profiles the user as novice. We refer to this as agent digression. In both there scenarios, the exchanges are constructed by using (node, edge) pair and FAQs related to the node.

%The number of exchanges in a digression depends on the user's level of expertise. When the user is proficient then they may ask specific questions about the process and the digression may last for a few exchanges. When the user is novice and has no clue about what the agent just said, then the digression can go on for a lot more exchanges. The maximum number of exchanges depends on the number of FAQs associated with the node. We manually annotated each node with the allowable permutations of associated FAQs that can make a meaningful digression. For each (node, edge) pair in the path, we toss a coin to decide if the outline should be appended with a single exchange or multiple exchanges. If we decide to use multiple exchanges, we then randomly choose the digression type (user/agent digression) and a valid permutation of FAQs associated with the node.

\vspace{0.5ex} 
\noindent \textbf{Closing:} Closing contains the solution suggested by the agent followed by one or more exchanges to gracefully terminate the dialog. Typically, the users thank the agent and the agent terminates the dialog by acknowledging the user.

\subsection{AMT Paraphrasing Tasks}

%We use this breakdown of dialogs, as described in \ref{fig:dialog_structure}, to construct the tasks for data collections using Mechanical Turk. Note that the block on agent-user utterance exchange is atomic and independent of the previous exchange. We exploit this observation to collect our dialog dataset as independent pairs of exchanges between agent and the user. We collect the agent questions as paraphrases of the flowchart nodes, and agent digressions as appropriate paraphrases of FAQ questions.\todo{insert a sample task} User utterances are then conditioned on these previously collected agent utterances to have context sensitive user responses. \todo{mention task instructions like uses of paraphrases, etc}  \todo{creation of tasks from the flowchart}

We crowdsource paraphrases of each utterance in a dialog outline. Utterances corresponding to each component (problem description, node-edge pairs in the flowchart path and closing) are paraphrased separately and then stitched together to construct a dialog. We define four types of paraphrasing tasks: non-contextual, contextual, problem description and closing tasks. In the non-contextual task, a single utterance from the outline is provided to the crowd workers to paraphrase. We requested the workers to provide two paraphrases for each utterance to improve diversity among paraphrases \cite{jiang-etal-2017-understanding,yaghoub-zadeh-fard-etal-2019-study}. In the contextual task, workers are asked to paraphrase in the context of a specific previously collected paraphrase. 
Problem descriptions tasks ask the worker to describe the troubleshooting problem using the primary issue and secondary issue as discussed in Section \ref{section:dialog_construction}. In closing task, the worker gracefully terminates the dialog in the context of a troubleshooting solution collected from a non-contextual task.
Examples of the four type of tasks can be seen in Figure \ref{fig:data-collection}b.

As most user responses in a flowchart are yes/no, we design the yes/no paraphrasing task based on a study by \newcite{RossenKnill1997YesNoQA}. 
%which explains how people typically respond to a yes/no question.  Inspired by the study, 
We add specific rules in the tasks for workers to follow when paraphrasing a yes/no user response. An example (in blue outline) is shown in Figure \ref{fig:data-collection}b.

\begin{comment}
\begin{figure}
   \includegraphics[width=0.95\linewidth]{contextual.png}
   \caption{Example user utterance paraphrasing task}
   \label{fig:amt}
\end{figure}
\end{comment}

%To make sure the collected data is of the highest quality, each of the collected dialog utterances was manually cleaned. After cleaning the collected data, the utterances for node-edge pairs and digressions are randomly sampled from the set of collected paraphrases and then stitched together into dialogs. The random sampling is possible because of our quantization of of a dialog. 

\subsection{Dialog Construction}

We generate around 110 outlines for each flowchart by equally dividing them amongst the paths in the flowchart. We generate a total of 1,369 outlines and then collect paraphrases of the constructed outline components. Finally the component paraphrases are stitched together to construct 1,369 dialogs as shown in Figure \ref{fig:data-collection}c. 

The paraphrases corresponding to an outline component are interchangeable across dialogs.
%with the paraphrases of the same outline component in a different dialog.
%The individual components in any given outline appears in other outlines. So the paraphrase (along with the contextual paraphrase) collected for such components are interchangeable without breaking the overall flow of the dialog.
We take advantage of this and generate an additional 1,369 dialogs by randomly interchanging paraphrases without breaking semantics. Our final set has 2,738 dialogs with an avg of 15.56 utterances per dialog. The agent and user utterances have an average of 14.95 and 16.17 words in them.

%Since the utterances are modelled as independent pairs, we can resample from within a dataset to create new dialog sets, making sure to preserve the dialog structure and only pick paraphrases from the corresponding data split. We use this idea to resample our dataset and double the number of dialogs. This split is summarized in Table \ref{tab:data_stats}. 

 %A pair of utterances between the agent and the user is defined as a turn, and our dataset has a total of 19686 turns. There are 7.19 turns over all the dialogs on average. When we tokenize the utterances by words, the average length of utterances is 15.56, the average length of agent utterances is 14.95 and 16.17 is the average length of user utterances.

\subsection{Paraphrase Cleaning} 
To avoid error propagation, we manually verify all paraphrases and correct errors in grammar, spelling and polarity. It took an author approximately 5 minutes per dialog for this step.
%\todo{M: recalculated. 20 hrs looked too much}
%non-contextual paraphrases before using them to collect contextual paraphrases. We then corrected the contextual paraphrases as well. We corrected errors in grammar, spelling, and polarity. 
An example of a polarity error is when the question `\textit{Do you have an open circuit?}' was paraphrased as `\textit{Do you have a closed circuit?}' by a crowd worker. Such paraphrases invert the semantics of (yes/no) edges from the given node and will break the correctness of a dialog, if not corrected. About 6\% utterances were recollected as they violated  instructions.

%In total, about 56\% of the collected paraphrases were corrected
%, 38\% didn't need any correction 
%and about 6\% were recollected as they violated the instructions. A histogram of the number of word edits \cite{levenshtein1966binary} between the collected and corrected paraphrase is shown in Figure \ref{fig:data_cleaning}.

%\begin{figure}[ht]
%\centering
%  \includegraphics[width=0.9\linewidth]{edit_distances.png}
% \caption{Histogram of word level Levenshtein distances between collected  and cleaned paraphrases}
%\label{fig:data_cleaning}
%\end{figure}

\section{Task Definition \& Baseline System}

In this section, we define the problem of learning flowchart grounded task oriented dialogs in an end-to-end manner without the use of intermediate labels. We then describe our proposed baseline model, \sys, which retrieves necessary knowledge from flowchart/FAQs and generates the agent response using the retrieved knowledge.

\subsection{Task Definition}

We represent a dialog $d$ between a user $u$ and an agent $a$ as a sequence of utterances $\{c_1^u, c_1^a,c_2^u, c_2^a,\ldots,c_m^u, c_m^a\}$, where $m$ denotes the number of exchanges in the dialog. %\todo{number of exchanges or number of turns?}. Each utterances is represented as a sequence of words. 
Let $\mathcal{F} = (N,E)$ be the flowchart over which the dialog $d$ is grounded, where the set of nodes $N$ represents the agent questions and edges $E$ represent the user responses. The number of outgoing edges from a node depends on the number of possible user responses for the agent question associated with the node. Let $\mathcal{Q}= \{q_l:a_l\}_{l=1}^{L}$ be the set of frequently asked question and answer pairs (FAQs) associated with the flowchart $\mathcal{F}$. Our objective is to learn a next response predictor, which takes (1) the dialog-history $\textbf{h}=\{c^u_1, c^a_1,\ldots,c^u_{i}\}$, (2) a flowchart ($\mathcal{F}$), and (3) a set of FAQs ($\mathcal{Q}$) as input and predicts the next agent response ($\mathbf{y}=c^a_i=\tuple{y_1 y_2 \ldots y_T}$).
%word by word. 
%\todo{we have not used $y$s anywhere. We should probably use $c^a_i$, SA:added y here} 

%The dialog system is trained in an end-to-end manner without the use of any explicit annotations on the dialog utterances.

%Let $\mathcal{F}$ be a given  flowchart for a given troubleshooting problem and let $\{q_i:a_i\}_{i=0}^{n_{faq}}$ be the set of accompanying FAQs ($n_{faq}$ being the number of FAQs for that flowchart). Let the dialog history so far be denoted by $\{u_j\}_{j=0}^{n_d}$ ($n_d$ is the number of utterances in the dialog so far). Then the objective of the \dataset\ challenge is to use the flowchart, accompanying FAQ document, and the dialog history together to generate the agent response $r$.
%$$r = F_{FLODIAL}(\{u_j\}_{j=0}^{n_d}, F,\{q_i:a_i\}_{i=0}^{n_{faq}})$$

\subsection{Baseline System: \sys}

Since it is a novel task, an existing TOD architecture does not directly apply on this problem. 
We design a baseline architecture named \sys\ for predicting the agent responses in flowchart grounded dialogs. \sys\ is trained in an end-to-end manner without the need for any intermediate annotations such as (a) whether the given agent utterance is grounded on a flowchart node or FAQs, or (b) the specific flowchart node or FAQ on which the agent utterance is grounded.

\sys\ follows the retrieval augmented generation framework (RAG) \cite{ragneurips20,Guu2020REALMRL}, which first retrieves the necessary knowledge to generate the response and then generates the agent response one word at a time by using the retrieved knowledge. The framework consists of two main components, a retriever and  a generator. 
%The retriever $p_\eta(z|\textbf{h})$ outputs a distribution over top-k documents based on the dialog history $\textbf{h}$. 
The retriever $p_\eta(z|\textbf{h})$ outputs a distribution over all documents based on the dialog history $\textbf{h}$.
The flowchart $\mathcal{F}$ and FAQs $\mathcal{Q}$ are represented as documents (discussed further in Section \ref{sec:ret-doc}). The generator $p_\theta(y_t|\textbf{h},z,y_{1:t-1})$ generates the agent response $y$ word by word by using the dialog history $\textbf{h}$ and a retrieved document $z$. We generate the response using RAG-Sequence model:
\begin{align}
p(\textbf{y}|\textbf{h}) = \!\!\!\!\sum_{z\in N\cup\mathcal{Q}} \!\!p_\eta(z|\textbf{h})\prod_{t=1}^{T}p_\theta(y_t|\textbf{h},z,y_{1:t-1})
\label{eq:rag-seq}
\end{align}
%\todo{SJ: We are not defining y anywhere. We are using $c^a_i$. We should probably define it somewhere., SA:added $\mathbf{y}=c^a_i$ in task definition. is that okay now?}
The overall network is trained by minimizing the negative log-likelihood of the response given by Equation \ref{eq:rag-seq}.
Following \newcite{ragneurips20}, we marginalize over all the documents using a top-k approximation. We use top-5 documents in our training implementation due to memory constraints. During inference, only the top-1 document is used because the dialog's agent responses need to be grounded on only one flowchart node or FAQ. This is unlike RAG where multiple documents extracted from Wikipedia can contribute to the expected output. See Appendix \ref{appendix:topk} for further details.
%Following \newcite{ragneurips20}, we also use just the top-k documents from the retriever to generate the response. \todo{Explain inference and describe why we don't use thorough decoding}.

\subsubsection{Retrievable Documents}
\label{sec:ret-doc}
The retrievable document set includes all  flowchart nodes and all FAQ QA pairs associated with the flowchart. In the original RAG model, each (Wikipedia) document had a single dense embedding, based on which a document was retrieved and used. However, for our setting, the content of a flowchart node will typically not be explicitly mentioned in the dialog history. Instead, the right node is best determined based on the flowchart structure -- the path to that node -- as expressed in the dialog history. Similarly, for FAQs,  a QA-pair will typically be matched on the question and the answer will be used in subsequent dialog.
%The set of retrievable documents in RAG is a dense vector index of Wikipedia. Constructing a similar index from flowchart and the FAQs will miss out on the structural similarity between a dialog and a flowchart path. 
%The dialog progresses along a path on the flowchart and the agent asks questions corresponding to the nodes on the path. 
%Dialogs also share structure with FAQs when the user asks a question and the agent replies using the corresponding FAQ answer.

Consequently, we represent each document as a key-value pair. The document-key is used by the retriever to compute $p_\eta(z|\textbf{h})$ and the document-value is used by the generator during response generation. We construct a document for each node in $\mathcal{F}$ and for each FAQ in $\mathcal{Q}$. The document-key of a flowchart node is the sequence of utterances corresponding to the nodes and edges in the path from the root. Its document-value is the agent utterance associated with it. For a FAQ, the document-key and value are the question and answer, respectively. 
%The set of retrievable documents consists of all nodes in the flowchart and all FAQs associated with the flowchart.

\subsubsection{Retriever \& Generator}
The retriever scores each document $z$ based on the dialog history. The dialog history is encoded using a hierarchical recurrent encoder \cite{DBLP:journals/corr/SordoniBVLSN15}. The encoder computes a dense representation of the history $\phi_h(\textbf{h})$. The document-key is also encoded using a hierarchical recurrent encoder to compute its vector representation $\phi_z(\textbf{z})$. For each document, we assign a score as negative of the Euclidean distance between  $\phi_h(\textbf{h})$ and $\phi_z(\textbf{z})$. 
The top-k scores are then passed through a Softmax layer to compute $p_\eta(z|\textbf{h})$. 
%\todo{Why we choose euclidean?}
We use GPT2 as the generator $p_\theta(\textbf{y}|\textbf{h},z)$ and it receives a separate input for each retrieved document $z$. The input to GPT2 is constructed by concatenating all the utterances in the dialog history along with the document-value. 
GPT2 input is described in detail in Appendix \ref{sec:gpt_input}. The response is decoded using beam search.

%The role of the generator is to output an appropriate agent response for the given dialog history $x$ and a retrieved template response $z_i$. Similar to REALM and RAG, we can use a pretrained model for our generator and finetune it for our task. For FloNet, we use GPT2 as our generator model. The construction of the GPT2 input is done exactly as in  \cite{DBLP:journals/corr/abs-1901-08149} and is described in Figure \ref{fig:construct_input} and it is fed to the model as in Figure \ref{fig:gpt_input}.

\begin{comment}
\begin{center}
  \includegraphics[width=\textwidth/2]{construct_input.pdf}
  \captionof{figure}{Construction of the input to the GPT2 language model}

\label{fig:construct_input}
\end{center}
\begin{center}
  \includegraphics[width=\textwidth/2]{gpt_input.pdf}
  \captionof{figure}{Example of input to the GPT2 language model}

\label{fig:gpt_input}
\end{center}
\end{comment}

\subsubsection{Pre-training}
\label{sec:pt}

To provide a good initialization to the retriever and the generator, we pre-train both the components separately. For each dialog history and response pair $(\textbf{h}, \textbf{y})$ in our dataset, we first identify the document over which the response is grounded using weak supervision \cite{zhao-etal-2020-knowledge-grounded}. The document whose document-value has the highest BLEU score \cite{papineni-etal-2002-bleu} w.r.t. the response $\textbf{y}$ is labeled as the pseudo grounded document.

The retriever is pre-trained using a contrastive loss \cite{1640964} by using the pseudo grounded document as the positive example and any other random document as a negative example. The generator is pre-trained by minimizing the negative log likelihood of the response given the dialog history and the document-value of the pseudo grounded document. Following \newcite{wolf2019transfertransfo}, we add a next-utterance classification loss to the negative log likelihood loss.
The classification loss is applied on the output of a linear classification layer which receives the last hidden state of the generator and outputs the probability of a given utterance being the correct agent response. We use randomly sampled incorrect utterances as negative examples to train the generator based classifier. 

\section{Experimental Setup \& Results}

%We first describe the data splits and evaluation metrics used for our task. We then discuss the results.

\begin{table}[t]
\centering
\footnotesize
\begin{tabular}{l c c}
\toprule
% remove the following columns: unique flowcharts, 
& \textbf{S-Flo} & \textbf{U-Flo}\\
\midrule
Train Dialogs & 1,470 & 1,470 \\
Val Dialogs & 374 & 374 \\
Test Dialogs & 484 & 498	\\
%No. of Flowcharts & 12 & 12 \\
%No. of Utterance & 39,372 & 39,372 \\
%Avg. Utterance Length & 15.56 & 15.56 \\
%& Val & 454 &	6,868  &	16.34\\
%& Test & 498 &	6,800  &	15.55\\
%\midrule
%Overall & & 2738 & 39,372  & 15.56\\
\bottomrule
\end{tabular}
\caption{Statistics of the dataset split.} %Average utterance length is the mean utterance length across a set calculated using the number of words in an utterance.} 
\label{tab:data_stats}
\end{table}

\subsection{Data Split} 
We create two different splits of the dialogs in \dataset. The \textit{S-Flo} split is used for evaluating the ability of \sys\ to generate responses by following flowchart and FAQs. The \textit{U-Flo} split is used to study the ability of \sys\ to generalize to flowcharts unseen during train in a zero-shot flowchart grounded response generation setting.

%To generate the \textit{S-Flo} split, we divided the dialogs associated with each flowchart as follows: 66\% for train set, 17\% for validation set and 17\% for test set. We randomly select a path in the flowchart and push all the dialogs that follow the path to one set. To generate the \textit{U-Flo} split, we group all dialogs associated with 8 flowcharts as train set, all dialogs from 2 flowcharts as validation set and the remaining 2 into test set. Thus, the U-Flo split has mutually exclusive sets of flowcharts in each set. Some statistics on the dataset split are shown in Table \ref{tab:data_stats}.

\newtexta{We first set aside the test dialogs for the \textit{S-Flo} and the \textit{U-Flo} split. The remaining dialogs are then divided into train and validation sets. This ensures a hidden test set for both the splits.The \textit{U-Flo} test set is constructed using all the dialogs associated with 2 randomly sampled flowcharts. To create the \textit{S-Flo} test set, we sampled 17\% of the dialogs associated with each flowchart excluding the 2 flowcharts used for the U-Flo test set. We create the train and validation sets using the remaining dialogs that are not a part of both the test sets. To generate the \textit{S-Flo} train and validation sets, we divided the remaining dialogs associated with each flowchart (excluding the 2 flowcharts in U-Flo test set) as follows: 66\% for train set and 17\% for validation set. To generate the \textit{U-Flo} split, we group dialogs associated with 8 flowcharts as train set and dialogs from 2 flowcharts as validation set. The U-Flo split has mutually exclusive sets of flowcharts in each set. Some statistics on the dataset split are shown in Table \ref{tab:data_stats}.}

\newtexta{The split mentioned above is different from the one used in the published EMNLP version. The split described in the published version does not have a hidden test set whereas the split mentioned above ensures a hidden test set. For completeness, we report the published numbers in Appendix \ref{ssec:old-results}}

\begin{table}[t]
\centering
\footnotesize
%\begin{tabular}{p{2.9cm} p{0.6cm} p{0.5cm} p{0.6cm} p{0.5cm}}
\begin{tabular}{l|rr|rr}
\toprule
\multirow{2}{*}{\textbf{Model}} & \multicolumn{2}{c|}{\textbf{S-Flo}} & \multicolumn{2}{c}{\textbf{U-Flo}} \\
\cmidrule{2-5}
& \text{\textbf{BLEU}} & \text{\textbf{PPL}} & \text{\textbf{BLEU}} & \text{\textbf{PPL}}\\
\midrule
%TF-IDF + GPT2 & 10.39 & 12.88 & 3.00 & 16.38 \\
%\sys\ Cascaded &14.47&9.34&7.26&11.97\\
%\sys\ (No PT) & 16.05 & 3.86 & 12.46 & 5.35\\
%\sys & 16.34 & 3.79 & 13.43 & 4.94\\

TF-IDF + GPT2 & 7.90 & 13.28 & 6.90 & 18.53 \\
\sys\ (No PT) & 16.66 & 4.19 & 13.01 & 5.99\\
\sys & \multicolumn{1}{r}{\textbf{19.89}} & \multicolumn{1}{r|}{\textbf{4.17}} & \multicolumn{1}{r}{\textbf{14.83}} & \multicolumn{1}{r}{\textbf{5.67}}\\
\midrule
\newtext{Oracle Ret. + GPT2} & 21.96 & \text{-} & 21.56 & \text{-} \\
\bottomrule
\end{tabular}
\caption{Next response prediction performance.} 
\label{tab:gen-performance}
\end{table}

\subsection{Evaluation Metrics}
We measure the ability to generate responses using two standard metrics: BLEU and perplexity. As \dataset\ contains the labels of the document (flowchart node or FAQ) over which each agent response is grounded on, we use recall@1 (R@1) to measure the retriever performance. \newtext{We also compute a task-specific metric called \textit{success rate} (SR) which is measured as the fraction of dialogs for which an algorithm retrieved the correct flowchart-node/FAQ for all the agent utterances in the dialog.}

\subsection{Implementation Details}
The models were implemented using PyTorch \cite{NEURIPS2019_9015}. We identify hyper-parameters using a grid-search and identified the best hyper-parameters based on the evaluation of the held-out validation sets. Each hyper-parameter combination was run ten times. We sample word embedding size from \{$50$, $100, 200, 300\}$, retriever learning rates ($lr_R$)\footnote{$a$E$b$ denotes $a\times 10^{b}$} from $\{$1E{-2}, 5E{-3}, 1E{-3}, 5E{-4}, 1E{-4}, 5E{-5}, 1E{-5}$\}$, generator learning rates ($lr_G$) from $\{$6.25E{-4}, 2.5E{-4}, 6.25E{-5}, 2.5E{-5}, 6.25E{-6},$ $ 2.5E{-6}, 6.5E{-7}, 2.5E{-7}$\}$, and dropout from increments of 0.02 between [0, 0.2]. Hidden size of the retriever was set to three times the word embedding size in all the settings. The word embeddings of the retriever were initialized with pre-trained GloVe embeddings \cite{pennington2014glove}. The generator was built on top code made available by \newcite{wolf2019transfertransfo}.\footnote{\url{https://github.com/huggingface/transfer-learning-conv-ai}}
The best hyper-parameter settings and other details are in Appendix \ref{ssec:training-details}

\subsection{Results}

\begin{table}[t]
\centering
\footnotesize
\begin{tabular}{l |c c |c c}
\toprule
\multirow{2}{*}{\textbf{Model}} & \multicolumn{2}{c|}{\textbf{S-Flo}} & \multicolumn{2}{c}{\textbf{U-Flo}}\\
\cmidrule{2-5}
&\textbf{R@1} & \textbf{SR} & \textbf{R@1} & \textbf{SR}\\
\midrule
%TF-IDF + GPT2 &0.334&0.836&0.394&0.877\\
%\sys\ (No PT) &0.768&0.944&0.586&0.878\\
%\sys &\textbf{0.814}&\textbf{0.957}&\textbf{0.661}&\textbf{0.916}\\

TF-IDF + GPT2 &0.304&0.002&0.373&0.004\\
\sys\ (No PT) &0.749&0.260&0.590&0.062\\
\sys &\textbf{0.793}&\textbf{0.318}&\textbf{0.677}&\textbf{0.133}\\
\bottomrule
\end{tabular}
\caption{Retriever performance of various models.} 
\label{tab:experiment_retriever}
\end{table}

%\todo{Rerun upper bound with oracle retriever, if the results are bad then remove the row}
We report the performance of our baseline \sys\ on both S-Flo and U-Flo splits of \dataset.
We also report the numbers for two simple variants of \sys: TF-IDF + GPT2 and \sys\ (No PT). The former variant uses a simple TF-IDF technique to retrieve documents. The top retrieved document concatenated with the dialog history is fed as input to GPT2 for generating a response.
%\todo{M: it looks weird that you did top-5 retrieved, and for TFIDF you give only top retrieved. SA: We use only top-1 for inference so it's the same} 
\sys\ (No PT) is \sys\ without the pre-training described in Section \ref{sec:pt}.

Table \ref{tab:gen-performance} reports the response prediction performance of various systems on both data splits and Table \ref{tab:experiment_retriever} reports the performance of the respective retrievers. TF-IDF + GPT2  has has a poor response prediction  performance in both the settings. The poor generalization is due to the TF-IDF retriever's low R@1.  This forces the generator to memorize the knowledge necessary to generate a response, rather than inferring it from the retrieved documents.

\sys\ achieves a three point improvement in BLEU over the No PT variant on S-Flo, and a two point jump in BLEU in U-Flo setting. This shows that the heuristic pre-training contributes to the overall system performance of \sys. \newtext{The success rate of various systems is reported in Table \ref{tab:experiment_retriever}. The success rate achieved by \sys{} retriever in both settings are quite low. We hope this gets improved by further research on the dataset.}

%\todo{D: do an experiemnt with top-k sampling SA:updated with both. M: where is it?} 
The oracle ret. + GPT2 in Table \ref{tab:gen-performance} is approximated by assuming a perfect retriever and training GPT2 with ground truth document. The gap in BLEU represents the value of annotation for our task, and the performance gain a better retriever may help achieve. 
%(flowchart node text/FAQ answer).
%The upper bound BLEU (Table \ref{tab:gen-performance}) is computed by comparing the gold response with the value of the document (flowchart node text/FAQ answer) over which the gold response is grounded.

We also compare the performance of \sys\ on the two data splits. We find that while numbers are understandably worse in the U-Flo setting, the zero-shot transferred \sys\ is still better than TF-IDF+GPT2's S-Flo performance. This suggests that the model has acquired some general intelligence of following a flowchart, even though there is significant scope for further improvement.

\begin{comment}
\begin{table}[t]
\centering
\footnotesize
%\begin{tabular}{p{2.9cm} p{0.6cm} p{0.5cm} p{0.6cm} p{0.5cm}}
\begin{tabular}{l|rr|rr}
\toprule
\multirow{2}{*}{\textbf{Model}} & \multicolumn{2}{c|}{\textbf{S-Flo}} & \multicolumn{2}{c}{\textbf{U-Flo}} \\
\cmidrule{2-5}
& \text{\textbf{Rel.}} & \text{\textbf{Gra.}} & \text{\textbf{Rel.}} & \text{\textbf{Gra.}}\\
\midrule
%TF-IDF + GPT2 & 10.39 & 12.88 & 3.00 & 16.38 \\
%\sys\ Cascaded &14.47&9.34&7.26&11.97\\
%\sys\ (No PT) & 16.05 & 3.86 & 12.46 & 5.35\\
%\sys & 16.34 & 3.79 & 13.43 & 4.94\\

%[(2.63, 1.13), (3.11, 2.37), (3.13, 2.55), (3.53, 3.69)]. % grammar scores are [(3.59, 3.24), (3.11, 3.62), (3.46, 2.71), (3.65, 3.76)]

TF-IDF + GPT2 & 2.63 & 3.59 & 1.13 & 3.24 \\
\sys\ (No PT) & 3.11 & 3.11 & 2.37 & \textbf{3.62}\\
\sys & \multicolumn{1}{r}{\textbf{3.12}} & \multicolumn{1}{r|}{\textbf{3.46}} & \multicolumn{1}{r}{\textbf{2.55}} & \multicolumn{1}{r}{2.71}\\
\midrule
Oracle Ret. + GPT2 & 3.53 & 3.65 & 3.69 & 3.76 \\

\bottomrule
\end{tabular}
\caption{\newtext{Human evaluation of various models.}} 
\label{tab:human-eval}
\end{table}
\end{comment}

\begin{table}[t]
\centering
\footnotesize
\begin{tabular}{l|SS|SS}
\toprule
\multirow{2}{*}{\textbf{Data Source}} & \multicolumn{2}{c|}{\textbf{S-Flo}} & \multicolumn{2}{c}{\textbf{U-Flo}}\\
\cmidrule{2-5}
& \text{\textbf{BLEU}} & \text{\textbf{PPL}} & \text{\textbf{BLEU}} & \text{\textbf{PPL}}\\
\midrule
%DH &10.24&14.64&2.51&19.40\\
%DH + FC &14.96&4.10&9.57&6.45\\
%DH + FC + FAQ & 16.34 & 3.79 & 13.43 & 4.94 \\
DH & \multicolumn{1}{r}{5.18} &\multicolumn{1}{r}{13.74}&\multicolumn{1}{r}{2.59}&\multicolumn{1}{r}{20.01}\\
DH + FC & \multicolumn{1}{r}{15.61} &\multicolumn{1}{r}{4.69}&\multicolumn{1}{r}{11.24}&\multicolumn{1}{r}{6.77}\\
DH + FC + FAQ & \multicolumn{1}{r}{\textbf{19.89}} & \multicolumn{1}{r|}{\textbf{4.17}} & \multicolumn{1}{r}{\textbf{14.83}} & \multicolumn{1}{r}{\textbf{5.67}} \\
\bottomrule
\end{tabular}
\caption{Response prediction performance of \sys\ with different knowledge sources. DH and FC indicates dialog history and flowchart respectively.}
\label{tab:experiment_data}
\end{table}

%\vspace{0.5ex} 
%\noindent \textbf{Human Evaluation:} 
%\newtext{We randomly sample 75 context-response pairs each from both S-Flo and U-Flo test sets and collect two sets of judgements for each pair. As we evaluate 4 systems, we collect a total of 1,200 labels from the judges. We report the human evaluation results in Table \ref{tab:human-eval}. We find that \sys’s relevance scores are better than the baselines for both S-Flo and U-Flo.}

\vspace{0.5ex} 
\noindent \textbf{Knowledge Sources:} To understand the contribution of each knowledge source towards response generation, we trained 3 variants of \sys: (i) using only the dialog history (\textit{DH}), (ii) using the dialog history and the flowchart (\textit{DH + FC}), and (iii) using dialog history, flowchart and FAQs (\textit{DH + FC + FAQ}). The performance is summarized in  Table \ref{tab:experiment_data}. 
% In indomain each knowledge source is important
The S-Flo trend shows both the knowledge sources contribute to the overall performance.
% Better generalization by learning from knowledge sources
The U-Flo numbers prove that, unsurprisingly, knowledge sources are essential for generalization to new settings, with more than 12 points increase in BLEU.
%As expected, the ability generalize to dialogs from the out-domain setting when no knowledge sources are used. By using the flowchart knowledge, the in-domain performance improves by 4 BLEU points and the system generalizes well on the out-domain setting. Finally, using both the knowledge sources achieves the best results. This shows that both the knowledge sources have to be leverage to achieve good in-domain performance and to bridge the gap between in-domain and out-domain performance.

%%%%Error analysis%%%%%
\section{Analysis \& Research Challenges}

\begin{table}[t]
\footnotesize
\begin{tabular}{l|c|c}
\toprule
\multicolumn{1}{c|}{\multirow{2}{*}{\textbf{Error Type}}} & \multicolumn{2}{c}{\textbf{\% Error (Count)}} \\
\cmidrule{2-3}
\multicolumn{1}{c}{} & \multicolumn{1}{|l|}{\textbf{S-Flo}} & \multicolumn{1}{l}{\textbf{U-Flo}} \\
\midrule
Retrieved Sibling & 68.2 (178) & 27.2 (193) \\
Retrieved Parent & 3.4 (9) & 7.6 (54) \\
Retrieved FAQ & 0.8 (2) & 2.1 (15) \\
Retrieved Other Nodes & 27.6 (72) & 63.0 (447) \\
\bottomrule
\end{tabular}
\caption{Retriever errors (\%) on utterances grounded on flowcharts. %The Error Type column shows the relationship of the retrieved document with the ground truth flowchart node. 
Error counts are in parentheses.
}
\label{tab:flowchart_ret_errors}
\end{table}

\begin{table}[t]
\footnotesize
\begin{tabular}{l|cc|cc}
\toprule
\multicolumn{1}{c}{\multirow{2}{*}{\textbf{Digression Type}}} & \multicolumn{2}{|c|}{\textbf{S-Flo}} & \multicolumn{2}{l}{\textbf{U-Flo}} \\
\cmidrule{2-5}
 & \textbf{BLEU} & \textbf{R@1} & \textbf{BLEU} & \textbf{R@1} \\
\midrule
%User Digression & 18.68 & 0.77 & 17.63 & 0.66 \\
%Agent Digression & 16.08 & 0.23 & 6.59 & 0.09 \\
User Digression & 29.27 & 0.88 & 7.84 & 0.53 \\
Agent Digression & 17.67 & 0.32 & 2.50 & 0.02 \\
\bottomrule
\end{tabular}
\caption{Performance of the generator (BLEU) and the retriever (R@1) on the utterances grounded on FAQs.}
\label{tab:faq-ret-errors}
\end{table}

We now investigate \sys\ errors in the validation set, with the goal of identifying new research challenges posed by \dataset. We first manually inspect the output of the generator, \emph{given} the retrieved document. We find that, by and large, the generator has learned its tasks well, which are deciding whether and how to use the retrieved document in generating a response. We attribute \sys's errors primarily to the retriever. 
%\todo{M: can you quickly analyze low BLEU examples when retriever is accurate. Maybe compare BLEU when retriever is correct vs when it is not? SA: checking} 
%This is also apparent from Recall@1 in Table \ref{tab:experiment_retriever}, which shows that 
\sys\ makes retrieval errors in 17.2\% and 46\% of validation examples in S-Flo and U-Flo, respectively.

To further diagnose retriever errors, we split them into two categories based on whether the correct retrieval is a flowchart node or a FAQ (digression). \newtext{For the former case, Table \ref{tab:flowchart_ret_errors} reports the nature of the error. In Table \ref{tab:flowchart_ret_errors}, retrieved sibling implies the retrieved node and the correct node are sibling nodes in the flowchart. We notice that for a large fraction of errors, retriever returns a sibling node. 
This  suggests that \sys\ could not adequately ground user response to the given agent question.} More surprising around 27\% of the errors in S-Flo are not even in the immediate neighborhood of the true node. A much larger value for U-Flo here also suggests poor retriever generalization. Since retriever performance in this task is closely tied with the ability to follow a flowchart path, it leads to the following research question: \emph{how can a model incorporate flowchart structure for better retriever performance?}

Table \ref{tab:faq-ret-errors} analyzes retrieval errors on digressions. We find that the retriever gets a decent Recall@1 for user digressions but has a rather low performance for agent digressions. Moreover, BLEU scores suggest that generator has memorized some common digressions S-Flo, but naturally they do not generalize to U-Flo. This yields a fairly challenging research question: \emph{how do we improve retriever performance on agent digressions?}

Finally, the challenge of zero-shot generalization to unseen flowcharts gets to the core ability of following a conversation flow, leading to the key research question: \emph{how do we improve performance on unseen flowchart setting in \dataset?}

\section{Conclusion}
We define the novel problem of end-to-end learning of flowchart grounded task oriented dialog (TOD) for a troubleshooting scenario. We collect a new flowchart grounded TOD dataset (\dataset), which contains 2,738 dialogs grounded on 12 different flowcharts and 138 FAQs. We propose the first baseline solution (\sys) for our novel task using retrieval-augmented generation. We outline novel technical challenges for TOD research identified in our work. We release \dataset\footnote{\url{https://dair-iitd.github.io/FloDial}} and all resources\footnote{\url{https://github.com/dair-iitd/FloNet}} 
for use by the research community.

\section*{Acknowledgments}
\label{sec:ack}
This work is supported by IBM AI Horizons Network grant, an IBM SUR award, grants by Google, Bloomberg and 1MG, a Visvesvaraya faculty award by Govt. of India, and the Jai Gupta chair fellowship by IIT Delhi. We thank Morris Rosenthal for providing us with permission to use the flowcharts from \url{www.ifitjams.com}. We also thank the IIT Delhi HPC facility for computational resources.

\section*{Ethics Impact Statement}
\vspace{0.5ex} 
\noindent \textbf{Crowd Worker Compensation:} The crowd workers were compensated with approximately 2.5 USD for a creating paraphrases for a dialog with 15 utterances. On an average, the crowd workers spent a little less than a minute on each paraphrase. Potentially, a worker can paraphrase 4 dialogs in an hour and get compensated with 10 USD.

\vspace{0.5ex} 
\noindent \textbf{Intellectual Property:} Flowcharts used in our data collection process are based on the flowcharts from \url{www.ifitjams.com}. We used these flowcharts after receiving a written permission from the creator Morris Rosenthal. We include attribution to Morris Rosenthal in the acknowledgements section.

\vspace{0.5ex} 
\noindent \textbf{Privacy:} 
%We systematically created 3 types of tasks to collect data from participants. 
We now briefly describe each task used for data collection and show how the task design ensures the collected data will not contain any sensitive personal information (SPI). We would like to emphasise that the authors of the paper meticulously went over each data point collected and removed the ones that did not comply with the rules in the task description.

\vspace{0.5ex} 
\noindent \textit{Problem Description Task:} Each participant was provided with an artificial scenario which includes a car/laptop model, a car/laptop model year, a car/laptop related problem that they are facing. They were requested to paraphrase this information into a natural language utterance. Since the scenario was provided by us, there was almost no room for providing SPI. The paraphrases deviating from the provided details were rejected.

\vspace{0.5ex} 
\noindent \textit{Paraphrasing Task:} The participants were requested to create paraphrases of a given sentence. This task has no room for providing SPI.

\vspace{0.5ex} 
\noindent \textit{Closing Task:}  The participants were asked to close a conversation between a human agent and a car/laptop user. In this task, the user and the human agent refer to each other using a second-person pronoun (e.g., I hope this was helpful to you, I am happy that this solved your problem). This task also does not involve providing any SPI.

% Entries for the entire Anthology, followed by custom entries
\bibliography{acl2021}

\begin{thebibliography}{40}
\expandafter\ifx\csname natexlab\endcsname\relax\def\natexlab#1{#1}\fi

\bibitem[{Bordes and Weston(2017)}]{BordesW16}
Antoine Bordes and Jason Weston. 2017.
\newblock Learning end-to-end goal-oriented dialog.
\newblock In \emph{International Conference on Learning Representations}.

\bibitem[{Budzianowski et~al.(2018)Budzianowski, Wen, Tseng, Casanueva, Ultes,
  Ramadan, and Gasic}]{budzianowski2018multiwoz}
Pawe{\l} Budzianowski, Tsung-Hsien Wen, Bo-Hsiang Tseng, I{\~n}igo Casanueva,
  Stefan Ultes, Osman Ramadan, and Milica Gasic. 2018.
\newblock Multiwoz-a large-scale multi-domain wizard-of-oz dataset for
  task-oriented dialogue modelling.
\newblock In \emph{Proceedings of the 2018 Conference on Empirical Methods in
  Natural Language Processing}, pages 5016--5026.

\bibitem[{Byrne et~al.(2019)Byrne, Krishnamoorthi, Sankar, Neelakantan,
  Goodrich, Duckworth, Yavuz, Dubey, Kim, and Cedilnik}]{byrne2019taskmaster}
Bill Byrne, Karthik Krishnamoorthi, Chinnadhurai Sankar, Arvind Neelakantan,
  Ben Goodrich, Daniel Duckworth, Semih Yavuz, Amit Dubey, Kyu-Young Kim, and
  Andy Cedilnik. 2019.
\newblock Taskmaster-1: Toward a realistic and diverse dialog dataset.
\newblock In \emph{Proceedings of the 2019 Conference on Empirical Methods in
  Natural Language Processing and the 9th International Joint Conference on
  Natural Language Processing (EMNLP-IJCNLP)}, pages 4506--4517.

\bibitem[{El~Asri et~al.(2017)El~Asri, Schulz, Sarma, Zumer, Harris, Fine,
  Mehrotra, and Suleman}]{el2017frames}
Layla El~Asri, Hannes Schulz, Shikhar~Kr Sarma, Jeremie Zumer, Justin Harris,
  Emery Fine, Rahul Mehrotra, and Kaheer Suleman. 2017.
\newblock Frames: a corpus for adding memory to goal-oriented dialogue systems.
\newblock In \emph{Proceedings of the 18th Annual SIGdial Meeting on Discourse
  and Dialogue}, pages 207--219.

\bibitem[{Eric et~al.(2017)Eric, Krishnan, Charette, and Manning}]{eric2017key}
Mihail Eric, Lakshmi Krishnan, Francois Charette, and Christopher~D Manning.
  2017.
\newblock Key-value retrieval networks for task-oriented dialogue.
\newblock In \emph{Proceedings of the 18th Annual SIGdial Meeting on Discourse
  and Dialogue}, pages 37--49.

\bibitem[{Feng et~al.(2020)Feng, Wan, Gunasekara, Patel, Joshi, and
  Lastras}]{feng2020doc2dial}
Song Feng, Hui Wan, Chulaka Gunasekara, Siva Patel, Sachindra Joshi, and Luis
  Lastras. 2020.
\newblock Doc2dial: A goal-oriented document-grounded dialogue dataset.
\newblock In \emph{Proceedings of the 2020 Conference on Empirical Methods in
  Natural Language Processing (EMNLP)}, pages 8118--8128.

\bibitem[{Gangi~Reddy et~al.(2019)Gangi~Reddy, Contractor, Raghu, and
  Joshi}]{gangi-reddy-etal-2019-multi}
Revanth Gangi~Reddy, Danish Contractor, Dinesh Raghu, and Sachindra Joshi.
  2019.
\newblock \href {https://doi.org/10.18653/v1/N19-1375} {Multi-level memory for
  task oriented dialogs}.
\newblock In \emph{Proceedings of the 2019 Conference of the North {A}merican
  Chapter of the Association for Computational Linguistics: Human Language
  Technologies, Volume 1 (Long and Short Papers)}, pages 3744--3754,
  Minneapolis, Minnesota. Association for Computational Linguistics.

\bibitem[{Guu et~al.(2020)Guu, Lee, Tung, Pasupat, and Chang}]{Guu2020REALMRL}
Kelvin Guu, Kenton Lee, Z.~Tung, Panupong Pasupat, and Ming-Wei Chang. 2020.
\newblock Realm: Retrieval-augmented language model pre-training.
\newblock \emph{ArXiv}, abs/2002.08909.

\bibitem[{Hadsell et~al.(2006)Hadsell, Chopra, and LeCun}]{1640964}
R.~Hadsell, S.~Chopra, and Y.~LeCun. 2006.
\newblock \href {https://doi.org/10.1109/CVPR.2006.100} {Dimensionality
  reduction by learning an invariant mapping}.
\newblock In \emph{2006 IEEE Computer Society Conference on Computer Vision and
  Pattern Recognition (CVPR'06)}, volume~2, pages 1735--1742.

\bibitem[{Ham et~al.(2020)Ham, Lee, Jang, and Kim}]{ham-etal-2020-end}
Donghoon Ham, Jeong-Gwan Lee, Youngsoo Jang, and Kee-Eung Kim. 2020.
\newblock \href {https://doi.org/10.18653/v1/2020.acl-main.54} {End-to-end
  neural pipeline for goal-oriented dialogue systems using {GPT}-2}.
\newblock In \emph{Proceedings of the 58th Annual Meeting of the Association
  for Computational Linguistics}, pages 583--592, Online. Association for
  Computational Linguistics.

\bibitem[{Henderson et~al.(2014)Henderson, Thomson, and Young}]{hen2014word}
Matthew Henderson, Blaise Thomson, and Steve Young. 2014.
\newblock Word-based dialog state tracking with re- current neural networks.
\newblock In \emph{In Proceedings of the 15th Annual Meeting of the Special
  Interest Group on Discourse and Dialogue (SIGDIAL)}, pages 292--299.

\bibitem[{Holtzman et~al.(2020)Holtzman, Buys, Du, Forbes, and
  Choi}]{Holtzman2020The}
Ari Holtzman, Jan Buys, Li~Du, Maxwell Forbes, and Yejin Choi. 2020.
\newblock \href {https://openreview.net/forum?id=rygGQyrFvH} {The curious case
  of neural text degeneration}.
\newblock In \emph{International Conference on Learning Representations}.

\bibitem[{Hosseini-Asl et~al.(2020)Hosseini-Asl, McCann, Wu, Yavuz, and
  Socher}]{NEURIPS2020_e9462095}
Ehsan Hosseini-Asl, Bryan McCann, Chien-Sheng Wu, Semih Yavuz, and Richard
  Socher. 2020.
\newblock \href
  {https://proceedings.neurips.cc/paper/2020/file/e946209592563be0f01c844ab2170f0c-Paper.pdf}
  {A simple language model for task-oriented dialogue}.
\newblock In \emph{Advances in Neural Information Processing Systems},
  volume~33, pages 20179--20191. Curran Associates, Inc.

\bibitem[{Jiang et~al.(2017)Jiang, Kummerfeld, and
  Lasecki}]{jiang-etal-2017-understanding}
Youxuan Jiang, Jonathan~K. Kummerfeld, and Walter~S. Lasecki. 2017.
\newblock \href {https://doi.org/10.18653/v1/P17-2017} {Understanding task
  design trade-offs in crowdsourced paraphrase collection}.
\newblock In \emph{Proceedings of the 55th Annual Meeting of the Association
  for Computational Linguistics (Volume 2: Short Papers)}, pages 103--109,
  Vancouver, Canada. Association for Computational Linguistics.

\bibitem[{Kelley(1984)}]{kelley1984iterative}
John~F Kelley. 1984.
\newblock An iterative design methodology for user-friendly natural language
  office information applications.
\newblock \emph{ACM Transactions on Information Systems (TOIS)}, 2(1):26--41.

\bibitem[{Kim et~al.(2020)Kim, Eric, Gopalakrishnan, Hedayatnia, Liu, and
  Hakkani-Tur}]{kim2020beyond}
Seokhwan Kim, Mihail Eric, Karthik Gopalakrishnan, Behnam Hedayatnia, Yang Liu,
  and Dilek Hakkani-Tur. 2020.
\newblock Beyond domain apis: Task-oriented conversational modeling with
  unstructured knowledge access.
\newblock In \emph{Proceedings of the 21th Annual Meeting of the Special
  Interest Group on Discourse and Dialogue}, pages 278--289.

\bibitem[{Kingma and Ba(2014)}]{kingma2014adam}
Diederik~P Kingma and Jimmy Ba. 2014.
\newblock Adam: A method for stochastic optimization.
\newblock \emph{arXiv preprint arXiv:1412.6980}.

\bibitem[{Lewis et~al.(2020)Lewis, Perez, Piktus, Petroni, Karpukhin, Goyal,
  K\"{u}ttler, Lewis, Yih, Rockt\"{a}schel, Riedel, and Kiela}]{ragneurips20}
Patrick Lewis, Ethan Perez, Aleksandra Piktus, Fabio Petroni, Vladimir
  Karpukhin, Naman Goyal, Heinrich K\"{u}ttler, Mike Lewis, Wen-tau Yih, Tim
  Rockt\"{a}schel, Sebastian Riedel, and Douwe Kiela. 2020.
\newblock \href
  {https://proceedings.neurips.cc/paper/2020/file/6b493230205f780e1bc26945df7481e5-Paper.pdf}
  {Retrieval-augmented generation for knowledge-intensive nlp tasks}.
\newblock In \emph{Advances in Neural Information Processing Systems},
  volume~33, pages 9459--9474. Curran Associates, Inc.

\bibitem[{Likert(1932)}]{likert1932technique}
Rensis Likert. 1932.
\newblock A technique for the measurement of attitudes.
\newblock \emph{Archives of psychology}.

\bibitem[{Papineni et~al.(2002)Papineni, Roukos, Ward, and
  Zhu}]{papineni-etal-2002-bleu}
Kishore Papineni, Salim Roukos, Todd Ward, and Wei-Jing Zhu. 2002.
\newblock \href {https://doi.org/10.3115/1073083.1073135} {{B}leu: a method for
  automatic evaluation of machine translation}.
\newblock In \emph{Proceedings of the 40th Annual Meeting of the Association
  for Computational Linguistics}, pages 311--318, Philadelphia, Pennsylvania,
  USA. Association for Computational Linguistics.

\bibitem[{Paszke et~al.(2019)Paszke, Gross, Massa, Lerer, Bradbury, Chanan,
  Killeen, Lin, Gimelshein, Antiga, Desmaison, Kopf, Yang, DeVito, Raison,
  Tejani, Chilamkurthy, Steiner, Fang, Bai, and Chintala}]{NEURIPS2019_9015}
Adam Paszke, Sam Gross, Francisco Massa, Adam Lerer, James Bradbury, Gregory
  Chanan, Trevor Killeen, Zeming Lin, Natalia Gimelshein, Luca Antiga, Alban
  Desmaison, Andreas Kopf, Edward Yang, Zachary DeVito, Martin Raison, Alykhan
  Tejani, Sasank Chilamkurthy, Benoit Steiner, Lu~Fang, Junjie Bai, and Soumith
  Chintala. 2019.
\newblock \href
  {http://papers.neurips.cc/paper/9015-pytorch-an-imperative-style-high-performance-deep-learning-library.pdf}
  {Pytorch: An imperative style, high-performance deep learning library}.
\newblock In H.~Wallach, H.~Larochelle, A.~Beygelzimer, F.~d\textquotesingle
  Alch\'{e}-Buc, E.~Fox, and R.~Garnett, editors, \emph{Advances in Neural
  Information Processing Systems 32}, pages 8024--8035. Curran Associates, Inc.

\bibitem[{Pennington et~al.(2014)Pennington, Socher, and
  Manning}]{pennington2014glove}
Jeffrey Pennington, Richard Socher, and Christopher~D. Manning. 2014.
\newblock \href {http://www.aclweb.org/anthology/D14-1162} {Glove: Global
  vectors for word representation}.
\newblock In \emph{Empirical Methods in Natural Language Processing (EMNLP)},
  pages 1532--1543.

\bibitem[{Radford et~al.(2019)Radford, Wu, Child, Luan, Amodei, and
  Sutskever}]{radford2019language}
Alec Radford, Jeff Wu, Rewon Child, David Luan, Dario Amodei, and Ilya
  Sutskever. 2019.
\newblock Language models are unsupervised multitask learners.

\bibitem[{Raghu et~al.(2019)Raghu, Gupta, and
  {Mausam}}]{raghu-etal-2019-disentangling}
Dinesh Raghu, Nikhil Gupta, and {Mausam}. 2019.
\newblock \href {https://doi.org/10.18653/v1/N19-1126} {{D}isentangling
  {L}anguage and {K}nowledge in {T}ask-{O}riented {D}ialogs}.
\newblock In \emph{Proceedings of the 2019 Conference of the North {A}merican
  Chapter of the Association for Computational Linguistics: Human Language
  Technologies, Volume 1 (Long and Short Papers)}, pages 1239--1255,
  Minneapolis, Minnesota. Association for Computational Linguistics.

\bibitem[{Raghu et~al.(2021)Raghu, Jain, {Mausam}, and
  Joshi}]{raghu-etal-2021-constraint}
Dinesh Raghu, Atishya Jain, {Mausam}, and Sachindra Joshi. 2021.
\newblock \href {https://doi.org/10.18653/v1/2021.findings-acl.448} {Constraint
  based knowledge base distillation in end-to-end task oriented dialogs}.
\newblock In \emph{Findings of the Association for Computational Linguistics:
  ACL-IJCNLP 2021}, pages 5051--5061, Online. Association for Computational
  Linguistics.

\bibitem[{Rastogi et~al.(2020)Rastogi, Zang, Sunkara, Gupta, and
  Khaitan}]{rastogi2020towards}
Abhinav Rastogi, Xiaoxue Zang, Srinivas Sunkara, Raghav Gupta, and Pranav
  Khaitan. 2020.
\newblock Towards scalable multi-domain conversational agents: The
  schema-guided dialogue dataset.
\newblock In \emph{Proceedings of the AAAI Conference on Artificial
  Intelligence}, volume~34, pages 8689--8696.

\bibitem[{Rossen-Knill et~al.(1997)Rossen-Knill, Spejewski, Hockey, Isard, and
  Stone}]{RossenKnill1997YesNoQA}
Deborah Rossen-Knill, Beverly Spejewski, Beth~Ann Hockey, Stephen Isard, and
  Matthew Stone. 1997.
\newblock Yes/no questions and answers in the map task corpus.
\newblock Technical report, University of Pennsylvania, Institute for Research
  in Cognitive Science.

\bibitem[{Saeidi et~al.(2018)Saeidi, Bartolo, Lewis, Singh, Rockt{\"a}schel,
  Sheldon, Bouchard, and Riedel}]{saeidi2018interpretation}
Marzieh Saeidi, Max Bartolo, Patrick Lewis, Sameer Singh, Tim Rockt{\"a}schel,
  Mike Sheldon, Guillaume Bouchard, and Sebastian Riedel. 2018.
\newblock Interpretation of natural language rules in conversational machine
  reading.
\newblock In \emph{Proceedings of the 2018 Conference on Empirical Methods in
  Natural Language Processing}, pages 2087--2097.

\bibitem[{Serban et~al.(2016)Serban, Sordoni, Bengio, Courville, and
  Pineau}]{serban2016building}
Iulian~Vlad Serban, Alessandro Sordoni, Yoshua Bengio, Aaron~C Courville, and
  Joelle Pineau. 2016.
\newblock Building end-to-end dialogue systems using generative hierarchical
  neural network models.
\newblock In \emph{AAAI}, pages 3776--3784.

\bibitem[{Shah et~al.(2018)Shah, Hakkani-T{\"u}r, T{\"u}r, Rastogi, Bapna,
  Nayak, and Heck}]{shah2018building}
Pararth Shah, Dilek Hakkani-T{\"u}r, Gokhan T{\"u}r, Abhinav Rastogi, Ankur
  Bapna, Neha Nayak, and Larry Heck. 2018.
\newblock Building a conversational agent overnight with dialogue self-play.
\newblock \emph{arXiv preprint arXiv:1801.04871}.

\bibitem[{Sordoni et~al.(2015)Sordoni, Bengio, Vahabi, Lioma, Simonsen, and
  Nie}]{DBLP:journals/corr/SordoniBVLSN15}
Alessandro Sordoni, Yoshua Bengio, Hossein Vahabi, Christina Lioma, Jakob~Grue
  Simonsen, and Jian{-}Yun Nie. 2015.
\newblock \href {http://arxiv.org/abs/1507.02221} {A hierarchical recurrent
  encoder-decoder for generative context-aware query suggestion}.
\newblock \emph{CoRR}, abs/1507.02221.

\bibitem[{Vinyals and Le(2015)}]{vinyals2015neural}
Oriol Vinyals and Quoc Le. 2015.
\newblock A neural conversational model.
\newblock \emph{Proceedings of the International Conference on Machine
  Learning, Deep Learning Workshop.}

\bibitem[{Wen et~al.(2016)Wen, Gasic, Mrk\v{s}i\'{c}, Rojas~Barahona, Su,
  Ultes, Vandyke, and Young}]{wenEMNLP2016}
Tsung-Hsien Wen, Milica Gasic, Nikola Mrk\v{s}i\'{c}, Lina~M. Rojas~Barahona,
  Pei-Hao Su, Stefan Ultes, David Vandyke, and Steve Young. 2016.
\newblock \href {https://aclweb.org/anthology/D16-1233} {Conditional generation
  and snapshot learning in neural dialogue systems}.
\newblock In \emph{EMNLP}, pages 2153--2162, Austin, Texas. ACL.

\bibitem[{Williams and Young(2007)}]{williams2007partially}
Jason~D. Williams and Steve Young. 2007.
\newblock \href {https://doi.org/https://doi.org/10.1016/j.csl.2006.06.008}
  {Partially observable markov decision processes for spoken dialog systems}.
\newblock \emph{Computer Speech \& Language}, 21(2):393--422.

\bibitem[{Wolf et~al.(2019)Wolf, Sanh, Chaumond, and
  Delangue}]{wolf2019transfertransfo}
Thomas Wolf, Victor Sanh, Julien Chaumond, and Clement Delangue. 2019.
\newblock Transfertransfo: A transfer learning approach for neural network
  based conversational agents.
\newblock \emph{arXiv preprint arXiv:1901.08149}.

\bibitem[{Wu et~al.(2019)Wu, Madotto, Hosseini-Asl, Xiong, Socher, and
  Fung}]{wu2019transferable}
Chien-Sheng Wu, Andrea Madotto, Ehsan Hosseini-Asl, Caiming Xiong, Richard
  Socher, and Pascale Fung. 2019.
\newblock Transferable multi-domain state generator for task-oriented dialogue
  systems.
\newblock In \emph{Proceedings of the 57th Annual Meeting of the Association
  for Computational Linguistics}, pages 808--819.

\bibitem[{Yaghoub-Zadeh-Fard et~al.(2019)Yaghoub-Zadeh-Fard, Benatallah,
  Chai~Barukh, and Zamanirad}]{yaghoub-zadeh-fard-etal-2019-study}
Mohammad-Ali Yaghoub-Zadeh-Fard, Boualem Benatallah, Moshe Chai~Barukh, and
  Shayan Zamanirad. 2019.
\newblock \href {https://doi.org/10.18653/v1/N19-1026} {A study of incorrect
  paraphrases in crowdsourced user utterances}.
\newblock In \emph{Proceedings of the 2019 Conference of the North {A}merican
  Chapter of the Association for Computational Linguistics: Human Language
  Technologies, Volume 1 (Long and Short Papers)}, pages 295--306, Minneapolis,
  Minnesota. Association for Computational Linguistics.

\bibitem[{Zhang et~al.(2020)Zhang, Sun, Galley, Chen, Brockett, Gao, Gao, Liu,
  and Dolan}]{zhang2020dialogpt}
Yizhe Zhang, Siqi Sun, Michel Galley, Yen-Chun Chen, Chris Brockett, Xiang Gao,
  Jianfeng Gao, Jingjing Liu, and William~B Dolan. 2020.
\newblock Dialogpt: Large-scale generative pre-training for conversational
  response generation.
\newblock In \emph{Proceedings of the 58th Annual Meeting of the Association
  for Computational Linguistics: System Demonstrations}, pages 270--278.

\bibitem[{Zhao and Eskenazi(2018)}]{zhao2018zero}
Tiancheng Zhao and Maxine Eskenazi. 2018.
\newblock Zero-shot dialog generation with cross-domain latent actions.
\newblock In \emph{Proceedings of the 19th Annual SIGdial Meeting on Discourse
  and Dialogue}, pages 1--10.

\bibitem[{Zhao et~al.(2020)Zhao, Wu, Xu, Tao, Zhao, and
  Yan}]{zhao-etal-2020-knowledge-grounded}
Xueliang Zhao, Wei Wu, Can Xu, Chongyang Tao, Dongyan Zhao, and Rui Yan. 2020.
\newblock \href {https://doi.org/10.18653/v1/2020.emnlp-main.272}
  {Knowledge-grounded dialogue generation with pre-trained language models}.
\newblock In \emph{Proceedings of the 2020 Conference on Empirical Methods in
  Natural Language Processing (EMNLP)}, pages 3377--3390, Online. Association
  for Computational Linguistics.

\end{thebibliography}
\bibliographystyle{acl_natbib}

\clearpage
\appendix

\section{Appendix}

\subsection{Training Details}
\label{ssec:training-details}

%This search was performed twice and the maximum validation scores were 16.62 and 9.72 for in-domain and out-domain respectively. 

\sys\ is trained in two phases: pre-train and fine-tune. In the pre-train phase, we use recall@1 and BLEU as the early stop criteria for the retriever and generator respectively. The recall is computed using the weakly supervised labels. The hyper-parameters (embedding size, $lr_R$, $lr_G$, dropout) that achieved the best validation numbers in pre-train stage were (100, 1E{-4}, 6.25E{-5}, 0.01) and (200, 1E{-4}, 6.25E{-5}, 0) for S-Flo and U-Flo respectively. The best hyper-parameters that achieved the best BLEU in fine tune phase were (100, 1E{-4}, 2.5E{-6}, 0) and (200, 1E{-5}, 2.5E{-7}, 0)  for S-Flo and U-Flo respectively. The BLEU scores on held-out validation set were 20.91 and 10.82 for S-Flo and U-Flo respectively. We use AdamW optimizer \cite{kingma2014adam} for training and beam search for decoding with beam width of 5 and a maximum decode length set to 60 tokens.

All experiments were run on a single Nvidia V100 GPU with 32GB of memory. The S-Flo retriever, U-Flo retriever and generator have 3M, 23M and 117M trainable parameters respectively. Thus, \sys\ has a total of 120M trainable parameters for S-Flo and 140M for U-Flo. \sys\ has an average runtime of approximately 7 hours (80 mins per epoch) and 8 hours (82 mins per epoch) for S-Flo and U-Flo respectively.

\subsection{GPT2 Input}
\label{sec:gpt_input}
\begin{figure}[h]
\centering
\includegraphics[width=\linewidth]{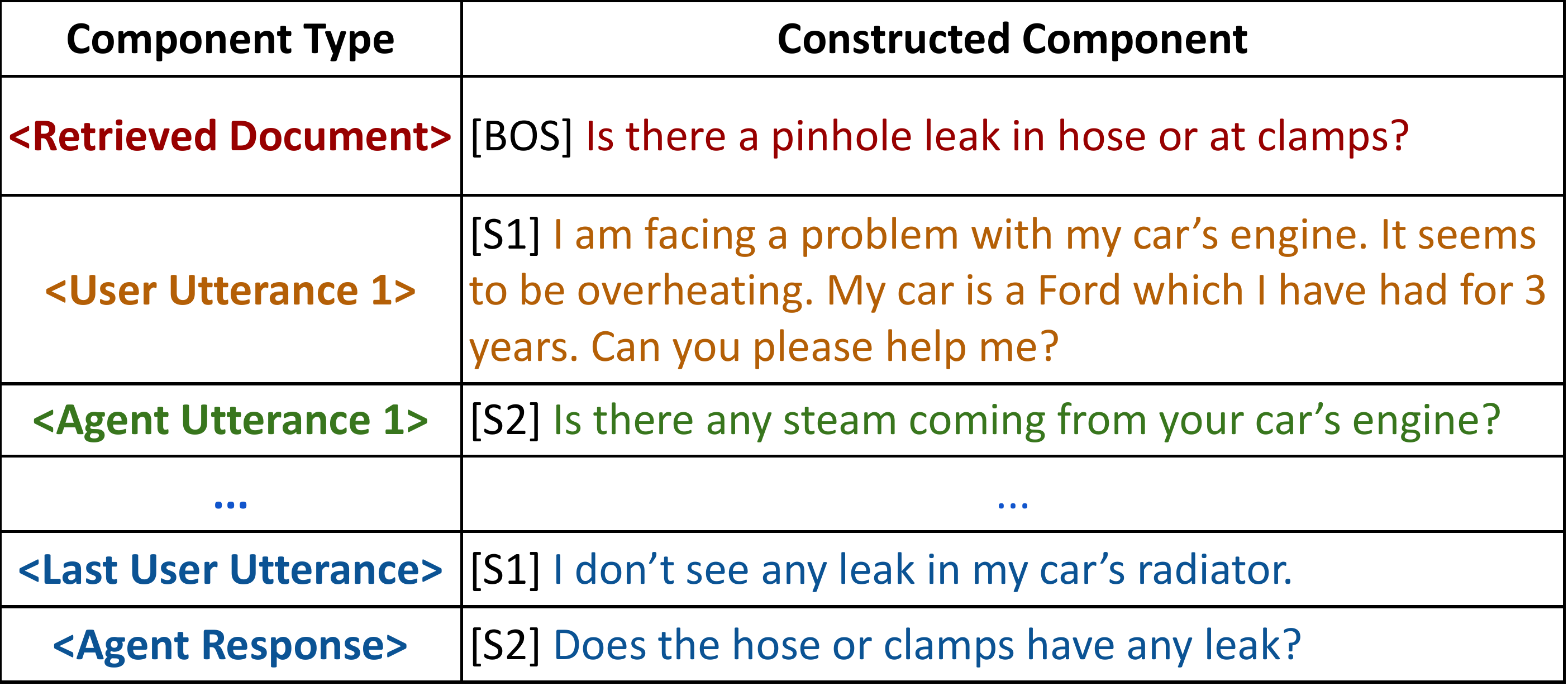}
\caption{Breakdown of the components in GPT2 input and output. %During training, we mask everything except \textit{<Agent Response>} in the expected output.
}
\label{fig:gpt_input}
\end{figure}

\sys\ uses GPT2 as the generator. Following \newcite{wolf2019transfertransfo}, our input is constructed as shown in Figure \ref{fig:gpt_input}. During inference, GPT2 take the retrieved document concatenated with the sequence of utterances from the dialog history as the input and predicts the agent response. %The target output is right shifted input with everything masked except the last agent response that we want to predict.

\subsection{GPT2 Inference}
\label{appendix:topk}
We experiment with various inference settings and used the setting which performed the best in the validation set. We experimented with beam search with beam width of 5. % two decoding techniques: nucleus sampling \cite{Holtzman2020The} with top-p as 0.9 and beam search with beam width of 5. 
We varied the number of top-k documents to be used. In the case of Top-1 decoding, we use only the top retrieved document to generate response candidates. For Top-5, we take the top 5 retrieved documents and generate candidate responses from each document. Each candidate response score is computed as a product of the probability of generating the candidate given the retrieved document  $\prod_{t=1}^{T}p_\theta(y_t|\textbf{h},z,y_{1:t-1})$ and the probability of the retrieved document $p_\eta(z|\textbf{h})$.
Lastly, we experimented with response length normalization to avoid favouring shorter sequences. The probability of each candidate is given by $(\prod_{t=1}^{T}p_\theta(y_t|\textbf{h},z,y_{1:t-1}))^{1/T}$ where $T$ is the length of the candidate. The validation BLEU scores of various settings on the S-Flo split is shown in table \ref{tab:sampling}. We see that beam search on top-1 document with length normalization resulted in the best validation BLEU.

%Table \ref{tab:sampling} shows that performance with Top-5 is inferior to Top-1. 
\begin{table}[!htbp]
\centering
\footnotesize
\begin{tabular}{l|c|c|c}
\toprule
\multirowcell{2}{\textbf{Decoding}\\\textbf{Technique}}&\multirow{2}{*}{\textbf{Top-k}}&\multirowcell{2}{\textbf{Length}\\\textbf{Norm.}}& \multirowcell{2}{\textbf{Val}\\ \textbf{BLEU}}\\
&&&\\
\midrule
\multirow{4}{*}{\textit{Beam}}&\multirow{2}{*}{Top-5}&No&18.22\\
&&Yes&20.08\\
\cmidrule{2-4}
&\multirow{2}{*}{\textit{Top-1}}&No&20.06\\
&&\textit{Yes}&\textbf{20.92} \\
\bottomrule
\end{tabular}
\caption{Validation and test BLEU scores of various settings on the S-Flo split.} 
\label{tab:sampling}
\end{table}

We also experimented with nucleus sampling \cite{Holtzman2020The} ($p= 0.9$). We found the validation BLEU was roughly three points lower than the BLEU achieved by beam search.

\subsection{Qualitative Examples} 
Table \ref{tab:gen-responses} and \ref{tab:gen-responses_in} shows responses generated by various systems on examples from U-Flo and S-Flo validation set respectively. In Table \ref{tab:gen-responses_in}, we see that \sys\ and \sys\ (No PT) generates responses similar to the gold response as they were able to generalize to unseen flowcharts. %\todo{Can we add an in-domain example where TF-IDF is incorrect. S:DONE}

\subsection{Results Reported in the EMNLP Version} 
\label{ssec:old-results}
\subsubsection{Data Split} 
To generate the S-Flo split, we divided the dialogs associated with each flowchart as follows: 66\% for train set, 17\% for validation set and 17\% for test set. We randomly select a path in the flowchart and push all the dialogs that follow the path to one set. To generate the U-Flo split, we group all dialogs associated with 8 flowcharts as train set, all dialogs from 2 flowcharts as validation set and the remaining 2 into test set. Thus, the U-Flo split has mutually exclusive sets of flowcharts in each set.

In this split, some dialogs in U-Flo test set are present in S-Flo train/validation set and some dialogs in S-Flo test set are present in U-Flo train/validation set. As this overlap of dialogs was preventing us from creating a hidden test set we defined a new split.

\subsubsection{Results} \newtexta{Table \ref{tab:old-gen-performance} reports the response prediction performance of various systems on both data splits and Table \ref{tab:old-experiment_retriever} reports the performance of the respective retrievers.}

\newtexta{We perform a human evaluation for the responses generated by \sys\ and 3 other variants of \sys\ along two dimensions: (i) \textit{relevance} -- the ability to generate responses that are relevant to the dialog context, and (ii) \textit{grammar} --  ability to generate grammatically correct and fluent responses. Both the dimensions are evaluated on a Likert scale (0-4) \cite{likert1932technique}.}

\begin{table}[t]
\centering
\footnotesize
%\begin{tabular}{p{2.9cm} p{0.6cm} p{0.5cm} p{0.6cm} p{0.5cm}}
\begin{tabular}{l|rr|rr}
\toprule
\multirow{2}{*}{\textbf{Model}} & \multicolumn{2}{c|}{\textbf{S-Flo}} & \multicolumn{2}{c}{\textbf{U-Flo}} \\
\cmidrule{2-5}
& \text{\textbf{BLEU}} & \text{\textbf{PPL}} & \text{\textbf{BLEU}} & \text{\textbf{PPL}}\\
\midrule
%TF-IDF + GPT2 & 10.39 & 12.88 & 3.00 & 16.38 \\
%\sys\ Cascaded &14.47&9.34&7.26&11.97\\
%\sys\ (No PT) & 16.05 & 3.86 & 12.46 & 5.35\\
%\sys & 16.34 & 3.79 & 13.43 & 4.94\\

TF-IDF + GPT2 & 11.97 & 12.88 & 6.45 & 16.38 \\
\sys\ (No PT) & 18.90 & 3.86 & 14.19 & 5.35\\
\sys & \multicolumn{1}{r}{\textbf{19.46}} & \multicolumn{1}{r|}{\textbf{3.79}} & \multicolumn{1}{r}{\textbf{16.31}} & \multicolumn{1}{r}{\textbf{4.94}}\\
\midrule
\newtext{Oracle Ret. + GPT2} & 23.73 & \text{-} & 24.85 & \text{-} \\
\bottomrule
\end{tabular}
\caption{Next response prediction performance. (From EMNLP Version)} 
\label{tab:old-gen-performance}
\end{table}

\begin{table}[t]
\centering
\footnotesize
\begin{tabular}{l |c c |c c}
\toprule
\multirow{2}{*}{\textbf{Model}} & \multicolumn{2}{c|}{\textbf{S-Flo}} & \multicolumn{2}{c}{\textbf{U-Flo}}\\
\cmidrule{2-5}
&\textbf{R@1} & \textbf{SR} & \textbf{R@1} & \textbf{SR}\\
\midrule
%TF-IDF + GPT2 &0.334&0.836&0.394&0.877\\
%\sys\ (No PT) &0.768&0.944&0.586&0.878\\
%\sys &\textbf{0.814}&\textbf{0.957}&\textbf{0.661}&\textbf{0.916}\\

TF-IDF + GPT2 &0.334&0.002&0.394&0.004\\
\sys\ (No PT) &0.768&0.260&0.586&0.064\\
\sys &\textbf{0.814}&\textbf{0.337}&\textbf{0.661}&\textbf{0.125}\\
\bottomrule
\end{tabular}
\caption{Retriever performance of various models. (From EMNLP Version)} 
\label{tab:old-experiment_retriever}
\end{table}

\begin{table}[t]
\centering
\footnotesize
%\begin{tabular}{p{2.9cm} p{0.6cm} p{0.5cm} p{0.6cm} p{0.5cm}}
\begin{tabular}{l|r|r}
\toprule
\multirow{2}{*}{\textbf{Model}} & \multicolumn{2}{c}{\textbf{Relevance}} \\
\cmidrule{2-3}
& \text{\textbf{S-Flo}} & \text{\textbf{U-Flo}} \\
\midrule
%TF-IDF + GPT2 & 10.39 & 12.88 & 3.00 & 16.38 \\
%\sys\ Cascaded &14.47&9.34&7.26&11.97\\
%\sys\ (No PT) & 16.05 & 3.86 & 12.46 & 5.35\\
%\sys & 16.34 & 3.79 & 13.43 & 4.94\\

%[(2.63, 1.13), (3.11, 2.37), (3.13, 2.55), (3.53, 3.69)]. % grammar scores are [(3.59, 3.24), (3.11, 3.62), (3.46, 2.71), (3.65, 3.76)]

TF-IDF + GPT2 & 2.63 & 1.13 \\
\sys\ (No PT) & 3.11  & 2.37 \\
\sys & \multicolumn{1}{r|}{\textbf{3.12}} & \multicolumn{1}{r}{\textbf{2.55}}\\
\midrule
Oracle Ret. + GPT2 & 3.53 & 3.69 \\

\bottomrule
\end{tabular}
\caption{\newtext{Human evaluation of various models.}} 
\label{tab:human-eval}
\end{table}

\newtexta{We randomly sample 75 context-response pairs each from both S-Flo and U-Flo test sets and collect two sets of judgements for each pair. As we evaluate 4 systems, we collect a total of 1,200 labels from the judges. We report the human evaluation results in Table \ref{tab:human-eval}. We find that \sys’s relevance scores are better than the baselines for both S-Flo and U-Flo.}

%\newtexta{Table \ref{tab:old-gen-performance} reports the generator performance on the test set, and table \ref{tab:old-experiment_retriever} reports the retriever performance.}

\subsection{Example Dialogs from \dataset}
Two randomly selected dialogs from \dataset are shown in Table \ref{tab:gen-dialogs}. The first dialog is grounded on the \textit{wireless network troubleshooting} flowchart
%, second is grounded on \textit{steering problems} flowchart 
and the second is grounded on \textit{car won't start} flowchart. 

\subsection{AMT Tasks}
\begin{figure*}[t]
\centering
\includegraphics[width=\textwidth]{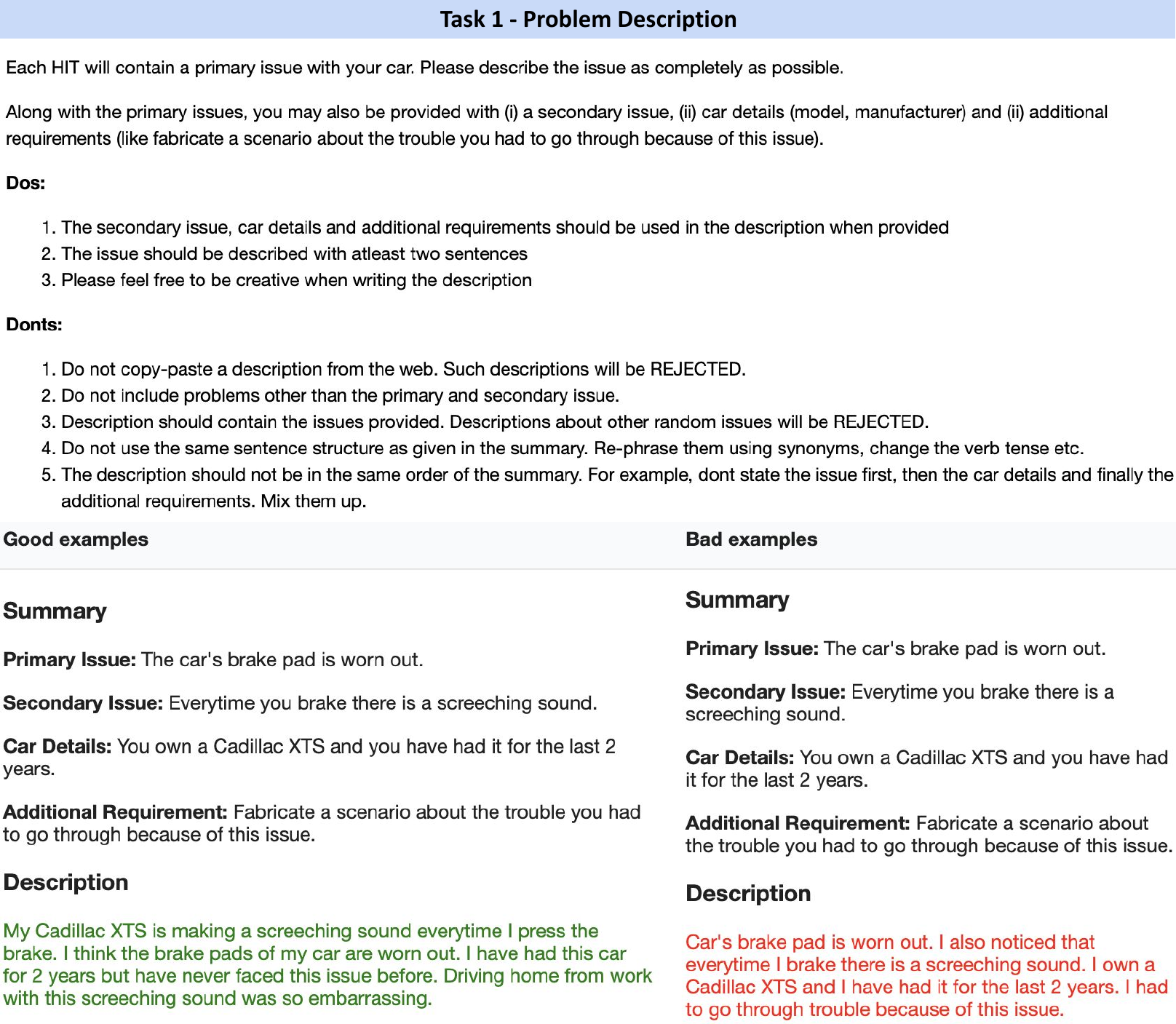}
\caption{Instructions provided to AMT workers for the problem description task.}
\label{fig:task1}
\end{figure*}

\begin{figure*}[t]
\centering
\includegraphics[width=\textwidth]{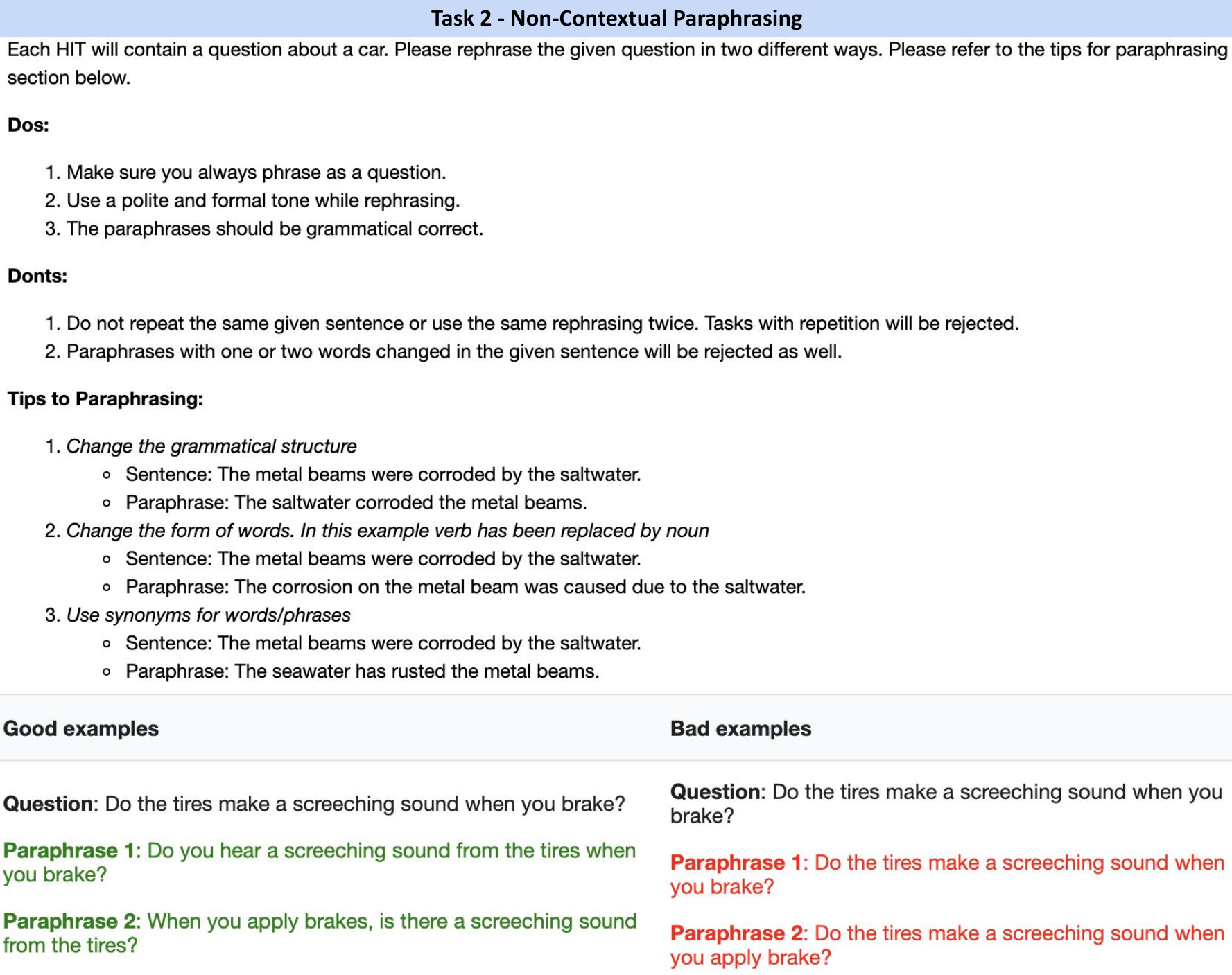}
\caption{Instructions provided to AMT workers for the non-contextual paraphrasing task.}
\label{fig:task2}
\end{figure*}

\begin{figure*}[ht]
\centering
\includegraphics[width=\textwidth]{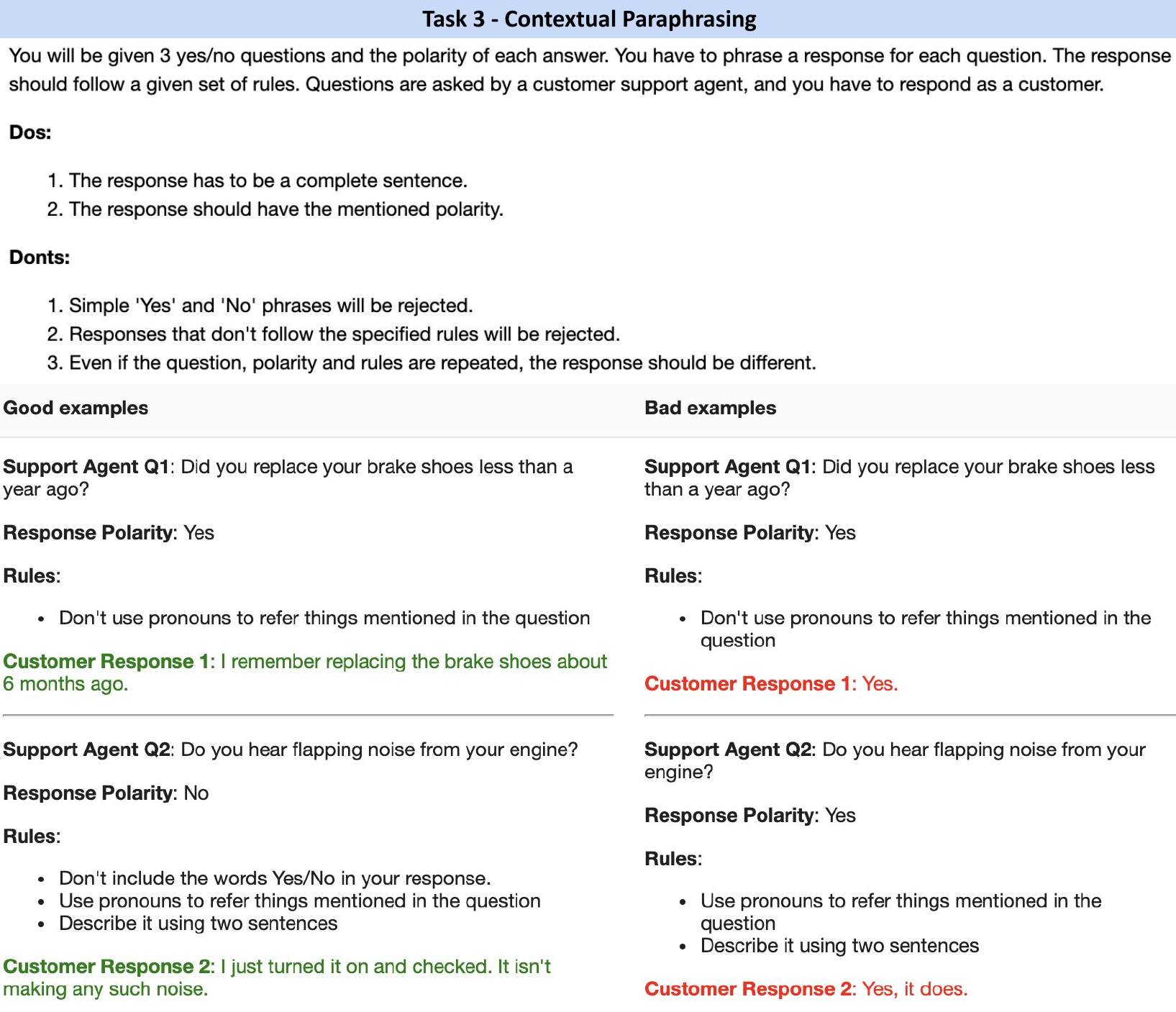}
\caption{Instructions provided to AMT workers for the contextual paraphrasing task.}
\label{fig:task3}
\end{figure*}

\begin{figure*}[ht]
\centering
\includegraphics[width=\textwidth]{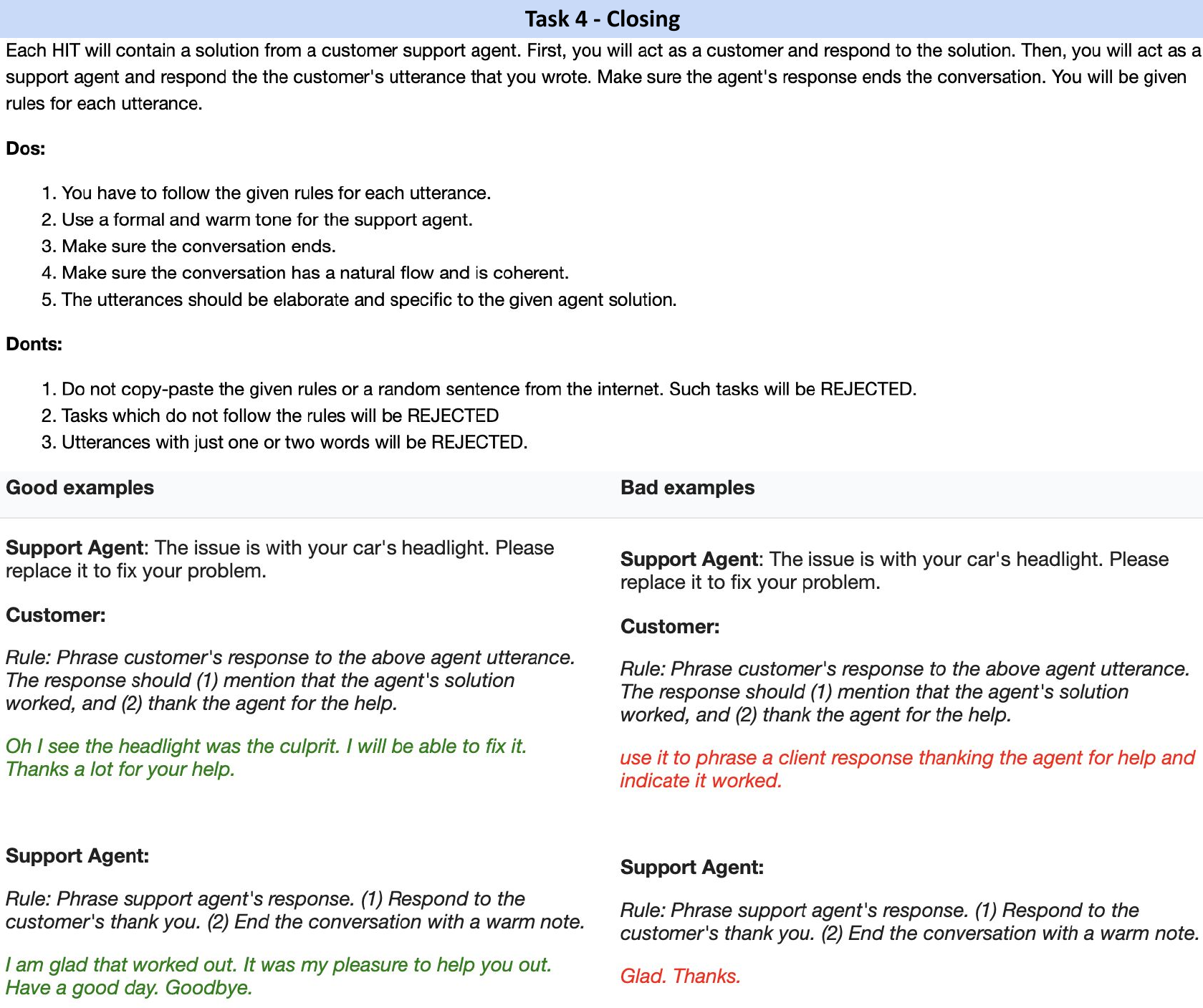}
\caption{Instructions provided to AMT workers for the closing task.}
\label{fig:task6}
\end{figure*}

Figure \ref{fig:task1}, \ref{fig:task2}, \ref{fig:task3} and \ref{fig:task6} show the instructions and examples provided to crowd-workers for problem description, non-contextual paraphrasing, contextual paraphrasing and closing task respectively.

\subsection{FAQ Collection}
\label{appendix:FAQ}

For each flowchart we collected a set of FAQs associated with it. We collect the FAQs in two steps: question collection and answer collection. In the question collection step, we displayed each flowchart node question to novice (laptop or car) users and asked them if they understand the node question. If not, they were requested to come up with a follow-up question that would help them better understand the node question. We aggregated all the follow-up questions to create the set of FAQs associated with that flowchart. In the answer collection step, in-house (laptop or car) expert provided answers for the questions collected in the first step.

%For collecting supporting FAQs for each flowchart, we asked in-house people, who are novice users of cars and laptops, if they are able to understand the information provided in the main flowchart. We collected FAQ questions from their responses and then asked in-house experts for answers to the questions.
%\subsection{Sample Output}
%Table \ref{tab:gen-responses} provides an example of response generated by the different baseline models discussed. The example is from out-domain test set.

\begin{table}[h]
\centering
\footnotesize
\begin{tabular}{l|p{4.5cm}}
\toprule
%\\
%Agent & Do you hear a clunking noise coming from your car?\\
\textit{User} & I've had my Infiniti QX60 for 6 years now. It's recently been making awful noises. I got stopped by the police on my way to work because the noise was no loud. Please help me fix this.\\
\midrule
\textit{Agent} & The sound that is coming from your car, is it a clunking sound?\\
\midrule
\textit{User} & I do hear a clunking sound coming from the car.\\
\midrule
\midrule
\textit{Gold Response} &Is it \hlc[lime!15]{only when} you \hlc[cyan!10]{drive over a bump or a dip} that the car makes a noise?\\
%\midrule
\midrule
TF-IDF + GPT2 &Are the ticking noises heard \hlc[lime!15]{only when} the car is moving?\\
\midrule
\sys\ (No PT) & Does the car make a noise \hlc[lime!15]{only when} you \hlc[cyan!10]{drive over bumps or dips} on the road?\\
\midrule
\sys&Does the car make the noise \hlc[lime!15]{only when} you \hlc[cyan!10]{drive it over a bump or a dip} on the road?\\
\bottomrule
\end{tabular}
\caption{Responses predicted by various algorithms for an example from the S-Flo validation set. The highlighted text shows that TF-IDF + GPT2 missed out on an important detail.} 
\label{tab:gen-responses_in}
\end{table}

\begin{table}[h]
\centering
\footnotesize
\begin{tabular}{l|p{4.5cm}}
\toprule
%\\
\textit{User} & My Dell G5 15 5590 is having some issues. I got it about a year ago but within the last month the battery just refuses to work properly. I don't understand what is happening here. This happens even if I plug it in overnight.\\
\midrule
\textit{Agent}&When the charger isn't plugged in, can the laptop power up?\\
\midrule
\textit{User} &I can see that it can start up in that case.\\
\midrule
\textit{Agent} & Do you find that the battery in your laptop quickly drains of power?\\
\midrule
\textit{User} & The battery drains too fast.\\
\midrule
\textit{Agent} & There will be a manufacture date on the laptop's battery. Please check if it is older than a year and let me know.\\
\midrule
\textit{User} & How do I check that date on my battery? \\
\midrule
\midrule
%Reference Document & Please remove the battery and you will find a sticker on it with a multi-field part number. The fourth field in the part number corresponds to the manufacturing date.\\
\textit{Gold Response} & In order to locate the manufacturing date, you need to remove the battery and find a sticker on it with a \hlc[lime!15]{multi-field part number}. The \hlc[cyan!10]{fourth field} in the number identifies the date.\\
%\midrule
\midrule
TF-IDF + GPT2 &You can use a multimeter to measure the voltage of the AC power adapter. The voltage should be the same as the voltage printed on the adapter.I am glad I could help. Have\\
\midrule
\sys\ (No PT) & Remove the battery and you will find a sticker on it with a \hlc[lime!15]{multi-field part number}.\\
\midrule
\sys&Please remove the battery and you will find a sticker on it with a \hlc[lime!15]{multi-field part number}. The \hlc[cyan!10]{fourth field} in the part number corresponds to the manufacturing date.\\
\bottomrule
\end{tabular}
\caption{Responses predicted by various algorithms for an example from the U-Flo validation set. The highlighted text points out that \sys\ generated the best response compared to other models.} 
\label{tab:gen-responses}
\end{table}

%\clearpage
%\newpage
\begin{table*}[!ht]
\centering
\footnotesize
\begin{tabular}{l|p{0.8\textwidth}}
\toprule
%\\
User & Hi, my Rivian R1T that I have had for only 1 year and I am already facing some issue with the brakes. I can't use it to go to work and i have to call for a ride everytime. I need to fix this soon. \\Agent & When you make use of the brakes, does the car stop moving? \\User & The brakes are able to stop the car from moving. \\Agent & Is the emergency brake functioning like it should? \\User & You know, I think it isn't functioning like usual. \\Agent & Are you experiencing excessive dragging from one or more wheels? \\User & The wheel on the car turn with ease. \\Agent & Are the brakes soft until you pump them a few times? \\User & The brakes are not soft. They do not need to be pumped.  \\Agent & Are there any noises heard after hitting the brakes? \\User & There is a distinct noise when I hit them.  I hadn't noticed this until just recently when I used them. \\Agent & When you apply the brakes, do they make a squealing noise? \\User & No, there is no squealing noises when the brakes are applied. \\Agent & Do you hear any noises like a clunk after applying the brakes? \\User & I checked just now. I can hear a clunk after I apply the brakes. \\Agent & There's a loose bolt and this is causing the caliper to move too freely on the wheel assembly. Please tighten the bolt. This should fix your braking problem. \\User & I tightened the loose caliper bolt, and it completely fixed my braking issue. Thanks! \\Agent & Glad to help you. I hope you car runs smoothly now.\\ 
\bottomrule
\toprule
User & I work from home and I am trying to finish a project.  I have a deadline and I cannot get my work done.  For some reason, I can't connect to the internet with my wireless network.  Help me please. \\Agent & Can you see the wireless network you are attempting to connect to on your laptop? \\User & Yes, I can see it. \\Agent & Can you connect to the wifi after disabling the router's network security settings? \\User & In order to turn off security in my router, what do I do?  \\Agent & Turn off or disable the security on the wireless router by opening the router settings in your browser. \\User & How can I access router settings from my browser? \\Agent & Try typing '192.168.1.1' or '192.168.0.1' in the address bar and you should see the router settings page. There you can log in to your router. \\User & Laptop doesn't connect with the security disabled. \\Agent & Can the computer connect to the internet when you connect directly to the router using an Ethernet cable? \\User & It does not connect to the internet. It doesn't connect even with an Ethernet cable.  \\Agent & Do you know how to check your laptop's firewall settings? \\User & I've checked the firewall settings before. \\Agent & Are the firewall settings blocking the internet connection? \\User & The firewall settings are not blocking my internet connection. \\Agent & Is the live internet LED on your router turned ON? \\User & I checked the live internet LED on the router. It is turned on. \\Agent & You are unable to connect to the internet because the router is malfunctioning. \\User & Replacing my malfunctioning router solved the issue. Thanks for your help. \\Agent & You're welcome. Have a nice day.\\
\bottomrule
\toprule
User & My Ford Expedition has started to make strange ticking sounds and I need it fixed immediately.  I'm going on a date and it has to go perfectly, with no unpleasant car noises.  The car is one year old and it's already making these ticking sounds when it moves. \\Agent & does the car make ticking sounds when it is moving in neutral? \\User & The car does make a ticking sound when in neutral. \\Agent & Do you know if the frequency of the ticking noise decreases once you shift gears in your car? \\User & I don't hear the sound decreasing when I shift gears. \\Agent & Are you hearing the ticking sound only when the car turns? \\User & I hear it all the time, not just during turns. \\Agent & Did you get new tires recently? \\User & They haven't been changed recently.  They are a few years old. \\Agent & Are the hub caps on your tires loose? \\User & The hub caps on my tires are not loose. The hub caps are secured very tightly. \\Agent & Have you inspected the treads in your car tires recently? \\User & I inspected the treads in my car tires yesterday. They look fine to me. \\Agent & Is it only when you are driving at a slow speed that you hear the ticking sounds? \\User & I hear the ticking sounds from it at high speed too. \\Agent & Your problem seems to be a common warped rotor issue. You''ll need to examine both the rotor as well as the brake pads.  \\User & Okay, I'll have to look into those issues.  Thank you so much for your help.   \\Agent & I'm glad that I could help you today.  Have a good day.\\
\bottomrule
\end{tabular}
\caption{Sample dialogs from \dataset} 
\label{tab:gen-dialogs}
\end{table*}

\clearpage

\end{document}